\documentclass[11pt, a4paper, logo, twocolumn, copyright]{googledeepmind}

\pdftrailerid{redacted}

\makeatletter
\renewcommand\bibentry[1]{\nocite{#1}{\frenchspacing\@nameuse{BR@r@#1\@extra@b@citeb}}}
\makeatother

\usepackage{longtable}
\usepackage{dsfont}
\usepackage{gdm-colors}
\usepackage{multirow}
\usepackage{graphicx}
\usepackage{rotating}
\usepackage{geometry}
\usepackage{xcolor}
\usepackage{placeins}
\usepackage{tabularx,tabulary}
\newcolumntype{Y}{>{\centering\arraybackslash}X}
\usepackage{colortbl}
\usepackage{hhline}

\usepackage[numbers, sort&compress, square]{natbib}

\makeatletter
\DeclareRobustCommand\onedot{\futurelet\@let@token\@onedot}
\def\@onedot{\ifx\@let@token.\else.\null\fi}

\def\eg/{\emph{e.g}\onedot} \def\Eg/{\emph{E.g}\onedot}
\def\ie/{\emph{i.e}\onedot} \def\Ie/{\emph{I.e}\onedot}
\def\cf/{\emph{c.f}\onedot} \def\Cf/{\emph{C.f}\onedot}
\def\etc/{\emph{etc}\onedot} \def\vs/{\emph{vs}\onedot}
\def\wrt/{w.r.t\onedot} \def\dof/{d.o.f\onedot}
\def\etal/{\emph{et al}\onedot}

\newcommand{\arxiv}[1]{}
\makeatother

\newcolumntype{L}[1]{>{\raggedright\let\newline\\\arraybackslash\hspace{0pt}}m{#1}} 

\title{SigLIP 2: Multilingual Vision-Language Encoders with Improved Semantic Understanding, Localization, and Dense Features}

\newcommand{\titlerunning}{SigLIP 2: Multilingual Vision-Language Encoders with Improved Semantic Understanding, Localization, and Dense Features}

\correspondingauthor{tschannen@google.com}

\renewcommand{\today}{February 2025}

\author[*,$\dagger$]{Michael Tschannen}
\author[*]{Alexey Gritsenko}
\author[*]{Xiao Wang}
\author[*]{Muhammad Ferjad Naeem}
\author[*]{Ibrahim Alabdulmohsin}
\author[*]{\\Nikhil Parthasarathy}
\author[*,$\circ$]{Talfan Evans}
\author[*,$\circ$]{Lucas Beyer}
\author[ \hspace{-0.6ex}]{Ye Xia}
\author[ \hspace{-0.6ex}]{Basil Mustafa}
\author[$\circ$]{Olivier H\'enaff}
\author[ \hspace{-0.6ex}]{Jeremiah Harmsen}
\author[ \hspace{-0.6ex}]{Andreas Steiner}
\author[*,$\circ$,$\dagger$]{Xiaohua Zhai}

\affil[ \hspace{-0.7ex}]{Google DeepMind}
\affil[*]{Core contributor}
\affil[$\dagger$]{Project lead}
\affil[$\circ$]{Work done while at Google DeepMind}

\begin{abstract}
We introduce SigLIP~2, a family of new multilingual vision-language encoders that build on the success of the original SigLIP. In this second iteration, we extend the original image-text training objective with several prior, independently developed techniques into a unified recipe---this includes captioning-based pretraining, self-supervised losses (self-distillation, masked prediction) and online data curation. With these changes, SigLIP~2 models outperform their SigLIP counterparts at all model scales in core capabilities, including zero-shot classification, image-text retrieval, and transfer performance when extracting visual representations for Vision-Language Models (VLMs). Furthermore, the new training recipe leads to significant improvements on localization and dense prediction tasks. We also train variants which support multiple resolutions and preserve the input's native aspect ratio. Finally, we train on a more diverse data-mixture that includes de-biasing techniques, leading to much better multilingual understanding and improved fairness. To allow users to trade off inference cost with performance, we release model checkpoints at four sizes: ViT-B (86M), L (303M), So400m (400M), and g (1B). 
\end{abstract}

\begin{document}

\maketitle

\section{Introduction}

Contrastive image-text embedding models trained on billion-scale datasets, as pioneered by CLIP~\cite{clip} and ALIGN~\cite{align}, have become the mainstream approach for high-level, semantic understanding of visual data. These models enable fine-grained, zero-shot classification rivaling the quality of supervised methods and enable efficient text-to-image and image-to-text retrieval. Furthermore, they lead to excellent vision-language understanding capabilities when combined with Large Language Models (LLMs) to build Vision-Language Models (VLMs).

Developing on the success of CLIP, several improvements have been proposed such as re-captioning images~\cite{maninis2024tips}, adding image-only self-supervised losses~\cite{naeem2024silc, maninis2024tips}, and training with a small decoder for auxiliary tasks such as captioning and localization~\cite{yu2022coca, blip2, locca}. At the same time, several groups have released model checkpoints for the open-source community~\cite{clip, zhai2022lit, ilharco2021open, sun2023eva, fang2024dfn}. However, these releases do not include the full breadth of latest improvements into a single model, as they all relatively closely follow CLIP's original approach. Here, building on the SigLIP training recipe~\cite{siglip}, we incorporate several improvements from prior work and release a new family of open models\footnote{Model checkpoints are available at \\\href{https://github.com/google-research/big_vision/tree/main/big_vision/configs/proj/image_text/README_siglip2.md}{https://github.com/google-research/big\_vision/tree/main/ big\_vision/configs/proj/image\_text/README\_siglip2.md}} that both excel on CLIP's core capabilities—--zero-shot classification, retrieval, and feature extraction for
VLMs---and improve areas where vanilla CLIP-style models lag behind, including localization and extracting dense, semantic representations.

In summary, SigLIP~2 models provide the following:
\begin{itemize}
    \item Strong multilingual vision-language encoders: SigLIP~2 shows excellent performance on English-focused vision-language tasks while providing strong results on multilingual benchmarks with a single model. This enables use in a wide range of languages and cultural contexts.
    \item Dense features: We incorporate self-supervised losses as well as a decoder-based loss, which result in better dense features (e.g. for segmentation and depth estimation) and improve localization tasks (such as referring expression comprehension).
    \item Backward compatibility: SigLIP~2 is designed to be backward compatible with SigLIP by relying on the same architecture. This allows existing users to simply swap out the model weights and tokenizer (which is now multilingual) to get improvements on a wide range of tasks.
    \item Native aspect ratio and variable resolution: SigLIP~2 also includes a NaFlex variant, which supports multiple resolutions and preserves the native image aspect ratio. These models have the potential to improve aspect sensitive applications such as document understanding.
    \item Strong small models: SigLIP~2 further optimizes performance of smaller models (B/16 and B/32 models), by using techniques in distillation via active data curation.
\end{itemize}

In the next section we provide a detailed description of the SigLIP~2 training recipe. Sec.~\ref{sec:experiments} presents evaluations of SigLIP~2 models and baselines across a variety of tasks and benchmarks. Finally, Sec.~\ref{sec:related} gives a short overview of related work, and conclusions can be found in Sec.~\ref{sec:conclusion}.

\begin{figure}[t]
    \centering
    \includegraphics[width=\columnwidth,trim={0, 2cm, 0, 0}, clip]{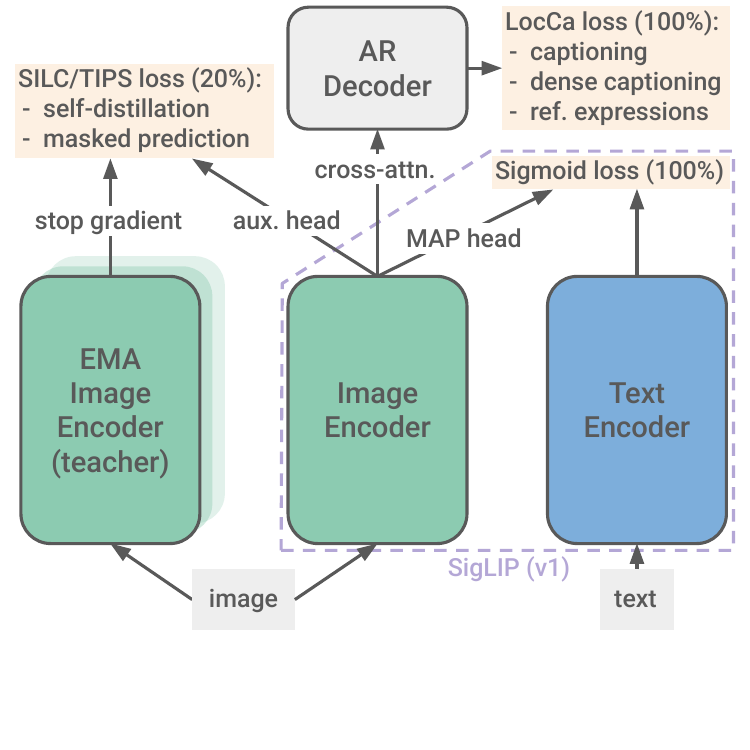}
    \caption{SigLIP 2 adds the captioning-based pretraining from LocCa~\cite{locca} as well as self-distillation and masked prediction from SILC~\cite{naeem2024silc} and TIPS~\cite{maninis2024tips} (during the last 20\% of training) to the sigmoid loss from SigLIP~\cite{siglip}. For some variants, the recipe additionally involves fine-tuning with data curation~\cite{udandarao2024active} or adaptation to native aspect ratio and variable sequence length~\cite{beyer2023flexivit, dehghani2024navit}.}
    \label{fig:overview}
\end{figure}

\section{Training recipe} \label{sec:training_recipe}

We combine the original SigLIP training recipe~\cite{siglip} with decoder-based pretraining~\cite{cappa, locca}, in addition to self-distillation and masked prediction as in the DINO line of work~\cite{caron2021emerging, oquab2024dinov2} (see Fig.~\ref{fig:overview} for an overview). Pretraining an image encoder with a language decoder for captioning and referring expression comprehension was shown to improve OCR capabilities and localization~\cite{locca}, whereas self-distillation and masked prediction leads to better features for dense prediction tasks, zero-shot classification and retrieval~\cite{naeem2024silc, maninis2024tips}. Rather than combining all these techniques in a single run we follow a staged approach as outlined below to manage the computational and memory overhead compared to SigLIP training.

In addition to training a set of models and adapting each model separately to different resolutions while distorting the aspect ratio, we also train variants which process images while largely preserving their native aspect ratio like NaViT~\cite{dehghani2024navit} and support different sequence lengths as FlexiViT~\cite{beyer2023flexivit}. We call this variant NaFlex, described in Sec.~\ref{sec:naflex}.

Finally, to improve the quality of the smallest models we fine-tune those with implicit distillation via active sample selection, following the approach from~\cite{udandarao2024active}.

\subsection{Architecture, training data, optimizer}
For the architecture, we follow SigLIP~\cite{siglip} so that existing users can simply swap out the encoder weights. Specifically, the fixed-resolution variant relies on the standard ViT architecture~\cite{dosovitskiy2021an} with learned positional embedding. We use the same architecture for the image and text tower, except for the g-sized vision encoder which is paired with an So400m-sized~\cite{sovit} text encoder. Vision and text representations are pooled using a MAP head (attention pooling)~\cite{scalingvit}. We set the text length to 64 and use the multilingual Gemma tokenizer~\cite{gemma} with vocabulary size 256k, transforming the text to lower case before tokenization.

We use the WebLI dataset \cite{pali} containing 10 billion images and 12 billion alt-texts covering 109 languages. To strike a good balance between quality on English and multilingual vision-language benchmarks we compose the mixture such that 90\% of the training image-text pairs is sourced from English web pages, and the remaining 10\% from non-English web pages, as recommended in~\cite{pouget2024no}. We further apply the filtering techniques from~\cite{alabdulmohsin2024clip} to mitigate data biases in representation and association with respect to sensitive attributes.

Unless noted otherwise, we use the Adam optimizer with learning rate $10^{-3}$, decoupled weight decay $10^{-4}$~\cite{loshchilov2017fixing}, and gradient clipping to norm 1. We set the batch size to 32k and use a cosine schedule with 20k warmup steps, training for a total of 40B examples. Our models are trained on up to 2048 TPUv5e chips~\cite{tpucloud} using a fully-sharded data-parallel strategy (FSDP~\cite{fsdp}).

\subsection{Training with Sigmoid loss and decoder}
\label{sec:siglip_training}

In the first step of pretraining, we combine SigLIP~\cite{siglip} with LocCa~\cite{locca} by simply combining the two losses with equal weight. Unlike CLIP~\cite{clip}, which relies on a contrastive loss, SigLIP creates binary classification problems by combining every image embedding with every text embedding in the mini-batch and trains the embeddings to classify matching and non-matching pairs via logistic regression (sigmoid loss). We use the original implementation and refer to~\cite{siglip} for details. 

For LocCa, we attach a standard transformer decoder with cross-attention to the un-pooled vision encoder representation (before applying the MAP head). The decoder follows the shapes of the text encoder except that we add cross-attention layers and reduce the number of layers by a factor of two. Besides image captioning, LocCa also trains for automatic referring expression prediction and grounded captioning. The former amounts to predicting bounding box coordinates for captions describing specific image regions, whereas the latter involves predicting region-specific captions given bounding box coordinates. Region-caption pairs are automatically annotated by first extracting n-grams from the alt-texts and then applying open-vocabulary detection using the recipe from~\cite{owlvitv2}. Additionally, we use the fixed set of object categories from~\cite{pali} instead of n-grams. For each example, the decoder is trained to predict all three targets (amounting to three decoder forward-passes). The captioning target is predicted with parallel prediction~\cite{cappa} with probability of 50\%, i.e. all caption tokens are predicted in parallel from mask tokens, without causal attention mask. Please refer to \cite{locca} for more detail. Finally, to reduce memory consumption due to the large vocabulary, we implement a chunked version of the decoder loss.

For all model sizes, we set the vision encoder patch size to 16 and the image resolution to 256 (resulting in an image representation sequence length of 256). Finally, we note that the decoder only serves for representation learning here and is not part of the model release.

\subsection{Training with self-distillation and masked prediction}
\label{sec:tips}

Following SILC~\cite{naeem2024silc} and TIPS~\cite{maninis2024tips}, we augment the training setup described in Sec.~\ref{sec:siglip_training} with local-to-global correspondence learning with self-distillation and masked prediction losses \cite{caron2021emerging, zhou2022image, oquab2024dinov2} to improve the local semantics of the (un-pooled) feature representation. This representation is typically used for dense prediction tasks like segmentation, depth estimation etc. Concretely, we add two terms to the losses described in Sec.~\ref{sec:siglip_training} as detailed next.

The first term is the local-to-global consistency loss from~\cite{naeem2024silc}, in which the vision encoder becomes the student network, which gets a partial (local) view of the training image, and is trained to match the teacher's representation, derived from the full image. This auxiliary matching task is performed in a high-dimensional feature space computed with a separate MLP head. As is common in the literature, the teacher parameters are obtained as an exponential moving average (EMA) of the student parameters over the previous iterations. We rely on a single global (teacher) view and 8 local (student) views and otherwise follow the augmentations, loss and hyper parameters from~\cite{naeem2024silc}.

The second loss term is the masked prediction objective from~\cite{maninis2024tips}. We replace 50\% of the embedded image patches in the student network with mask tokens and train the student to match the features of the teacher at masked locations. The loss is then defined identically to the first term (consistency loss), but applied to  per-patch features rather than the pooled, image-level representation. Further, both the student and the teacher see the same, global view (up to masking in the student).

We add these losses at 80\% of training completion, initializing the teacher with the student parameters and the remaining additional parameters (heads, mask token and corresponding optimizer parameters) randomly. We use the original image for computing the SigLIP and LocCa losses from the previous section and apply the additional losses on additional augmented views. This is done to ensure that data augmentation does not negatively impact the image-text alignment as recommended by ~\cite{naeem2024silc}. The weights of the first and the second loss terms are set to 1 and 0.25. Further, to balance model quality on global/semantic and dense tasks, we re-weight the two loss terms by another factor of 0.25, 0.5, 1.0, and 0.5 for the B, L, So400m and g, model sizes, respectively. 

\subsection{Adaptation to different resolutions}

\begin{table*}[t]
\vspace{-0.3cm}
\centering
\footnotesize
\setlength{\tabcolsep}{0.52em}
\begin{tabular}{lcclccccccccccc}
\toprule
 &  &  &  & \multicolumn{5}{c}{ImageNet-1k} & \multicolumn{2}{c}{COCO} & \multicolumn{2}{c}{Flickr} & \multicolumn{2}{c}{XM3600} \\ \cmidrule(lr){5-9} \cmidrule(lr){10-11} \cmidrule(lr){12-13} \cmidrule(lr){14-15}
ViT & Res. & Seq. & Model & val & v2 & ReaL & ObjNet & 10s. & T$\rightarrow$I & I$\rightarrow$T & T$\rightarrow$I & I$\rightarrow$T & T$\rightarrow$I & I$\rightarrow$T \\
\midrule
\multirow[c]{3}{*}{B/32} & 224 & 49 & MetaCLIP \cite{xu2024demystifying} & 67.7 & 59.6 & -- & 52.8 & -- & \underline{46.6} & -- & \underline{72.9} & -- & -- & -- \\
\arrayrulecolor{lightgray}\hhline{|~|--------------|} 
 & \multirow[c]{2}{*}{256} & \multirow[c]{2}{*}{64} & OpenCLIP \cite{ilharco2021open} & \underline{72.8} & \underline{64.8} & -- & \underline{59.6} & -- & 39.9 & \underline{57.9} & 64.9 & \underline{84.8} & -- & -- \\
 &  &  & \cellcolor{gray!15}SigLIP 2 & \cellcolor{gray!15}\bf{74.0} & \cellcolor{gray!15}\bf{66.9} & \cellcolor{gray!15}\bf{81.4} & \cellcolor{gray!15}\bf{66.1} & \cellcolor{gray!15}\bf{66.6} & \cellcolor{gray!15}\bf{47.2} & \cellcolor{gray!15}\bf{63.7} & \cellcolor{gray!15}\bf{75.5} & \cellcolor{gray!15}\bf{89.3} & \cellcolor{gray!15}\bf{38.3} & \cellcolor{gray!15}\bf{49.0} \\
\arrayrulecolor{black}\hhline{|---------------|} 
\multirow[c]{13}{*}{B/16} & \multirow[c]{7}{*}{224} & \multirow[c]{7}{*}{196} & CLIP \cite{clip} & 68.3 & 61.9 & -- & 55.3 & -- & 33.1 & 52.4 & 62.1 & 81.9 & -- & -- \\
 &  &  & OpenCLIP \cite{ilharco2021open} & 70.2 & 62.3 & -- & 56.0 & -- & 42.3 & 59.4 & 69.8 & 86.3 & -- & -- \\
 &  &  & MetaCLIP \cite{xu2024demystifying} & 72.4 & 65.1 & -- & 60.0 & -- & 48.9 & -- & 77.1 & -- & -- & -- \\
 &  &  & EVA-CLIP \cite{sun2023eva} & 74.7 & 67.0 & -- & 62.3 & -- & 42.2 & 58.7 & 71.2 & 85.7 & -- & -- \\
 &  &  & SigLIP \cite{siglip} & 76.2 & 69.5 & 82.8 & 70.7 & 69.9 & 47.2 & 64.5 & 77.9 & 89.6 & 22.4 & 29.3 \\
 &  &  & DFN \cite{fang2024dfn} & 76.2 & 68.2 & -- & 63.2 & -- & 51.9 & -- & 77.3 & -- & -- & -- \\
 &  &  & \cellcolor{gray!15}SigLIP 2 & \cellcolor{gray!15}78.2 & \cellcolor{gray!15}71.4 & \cellcolor{gray!15}84.8 & \cellcolor{gray!15}73.6 & \cellcolor{gray!15}72.1 & \cellcolor{gray!15}52.1 & \cellcolor{gray!15}68.9 & \cellcolor{gray!15}80.7 & \cellcolor{gray!15}93.0 & \cellcolor{gray!15}40.3 & \cellcolor{gray!15}50.7 \\
\arrayrulecolor{lightgray}\hhline{|~|--------------|} 
 & \multirow[c]{2}{*}{256} & \multirow[c]{2}{*}{256} & SigLIP \cite{siglip} & 76.7 & 70.1 & 83.1 & 71.3 & 70.3 & 47.4 & 65.1 & 78.3 & 91.1 & 22.5 & 29.9 \\
 &  &  & \cellcolor{gray!15}SigLIP 2 & \cellcolor{gray!15}79.1 & \cellcolor{gray!15}72.5 & \cellcolor{gray!15}85.4 & \cellcolor{gray!15}74.5 & \cellcolor{gray!15}73.1 & \cellcolor{gray!15}53.2 & \cellcolor{gray!15}69.7 & \cellcolor{gray!15}81.7 & \cellcolor{gray!15}94.4 & \cellcolor{gray!15}40.7 & \cellcolor{gray!15}51.0 \\
\arrayrulecolor{lightgray}\hhline{|~|--------------|} 
 & \multirow[c]{2}{*}{384} & \multirow[c]{2}{*}{576} & SigLIP \cite{siglip} & 78.6 & 72.0 & 84.6 & 73.8 & 72.7 & 49.7 & 67.5 & 80.7 & 92.2 & 23.3 & 30.3 \\
 &  &  & \cellcolor{gray!15}SigLIP 2 & \cellcolor{gray!15}\underline{80.6} & \cellcolor{gray!15}\underline{73.8} & \cellcolor{gray!15}\underline{86.2} & \cellcolor{gray!15}\underline{77.1} & \cellcolor{gray!15}\underline{74.7} & \cellcolor{gray!15}\underline{54.6} & \cellcolor{gray!15}\bf{71.4} & \cellcolor{gray!15}\underline{83.8} & \cellcolor{gray!15}\underline{94.9} & \cellcolor{gray!15}\underline{41.2} & \cellcolor{gray!15}\underline{51.6} \\
\arrayrulecolor{lightgray}\hhline{|~|--------------|} 
 & \multirow[c]{2}{*}{512} & \multirow[c]{2}{*}{1024} & SigLIP \cite{siglip} & 79.2 & 72.9 & 84.9 & 74.8 & 73.3 & 50.4 & 67.6 & 81.6 & 92.5 & 23.5 & 30.5 \\
 &  &  & \cellcolor{gray!15}SigLIP 2 & \cellcolor{gray!15}\bf{81.2} & \cellcolor{gray!15}\bf{74.5} & \cellcolor{gray!15}\bf{86.7} & \cellcolor{gray!15}\bf{77.8} & \cellcolor{gray!15}\bf{75.2} & \cellcolor{gray!15}\bf{55.2} & \cellcolor{gray!15}\underline{71.2} & \cellcolor{gray!15}\bf{84.5} & \cellcolor{gray!15}\bf{95.5} & \cellcolor{gray!15}\bf{41.4} & \cellcolor{gray!15}\bf{52.0} \\
\arrayrulecolor{black}\hhline{|---------------|} 
\multirow[c]{6}{*}{L/14} & \multirow[c]{6}{*}{224} & \multirow[c]{6}{*}{256} & OpenCLIP \cite{ilharco2021open} & 74.0 & 61.1 & -- & 66.4 & -- & 46.1 & 62.1 & 75.0 & 88.7 & -- & -- \\
 &  &  & CLIP \cite{clip} & 75.5 & 69.0 & -- & 69.9 & -- & 36.5 & 56.3 & 65.2 & 85.2 & -- & -- \\
 &  &  & MetaCLIP \cite{xu2024demystifying} & 79.2 & 72.6 & -- & 74.6 & -- & \underline{55.7} & -- & \underline{83.3} & -- & -- & -- \\
 &  &  & CLIPA-v2 \cite{li2023clipa} & 79.7 & 72.8 & -- & 71.1 & -- & 46.3 & \bf{64.1} & 73.0 & \underline{89.1} & -- & -- \\
 &  &  & EVA-CLIP \cite{sun2023eva} & \underline{79.8} & \underline{72.9} & -- & \bf{75.3} & -- & 47.5 & \underline{63.7} & 77.3 & \bf{89.7} & -- & -- \\
 &  &  & DFN \cite{fang2024dfn} & \bf{82.2} & \bf{75.7} & -- & \underline{74.8} & -- & \bf{59.6} & -- & \bf{84.7} & -- & -- & -- \\
\arrayrulecolor{black}\hhline{|---------------|} 
\multirow[c]{5}{*}{L/16} & \multirow[c]{2}{*}{256} & \multirow[c]{2}{*}{256} & SigLIP \cite{siglip} & 80.5 & 74.2 & 85.9 & 77.9 & 76.8 & 51.2 & 69.6 & 81.3 & 92.0 & 30.9 & 40.1 \\
 &  &  & \cellcolor{gray!15}SigLIP 2 & \cellcolor{gray!15}82.5 & \cellcolor{gray!15}76.8 & \cellcolor{gray!15}87.3 & \cellcolor{gray!15}83.0 & \cellcolor{gray!15}78.8 & \cellcolor{gray!15}54.7 & \cellcolor{gray!15}\underline{71.5} & \cellcolor{gray!15}84.1 & \cellcolor{gray!15}94.5 & \cellcolor{gray!15}46.5 & \cellcolor{gray!15}\underline{56.5} \\
\arrayrulecolor{lightgray}\hhline{|~|--------------|} 
 & \multirow[c]{2}{*}{384} & \multirow[c]{2}{*}{576} & SigLIP \cite{siglip} & 82.1 & 75.9 & 87.1 & 80.9 & 78.7 & 52.8 & 70.5 & 82.6 & 92.9 & 31.4 & 39.7 \\
 &  &  & \cellcolor{gray!15}SigLIP 2 & \cellcolor{gray!15}\underline{83.1} & \cellcolor{gray!15}\underline{77.4} & \cellcolor{gray!15}\underline{87.6} & \cellcolor{gray!15}\underline{84.4} & \cellcolor{gray!15}\underline{79.5} & \cellcolor{gray!15}\bf{55.3} & \cellcolor{gray!15}71.4 & \cellcolor{gray!15}\underline{85.0} & \cellcolor{gray!15}\underline{95.2} & \cellcolor{gray!15}\underline{47.1} & \cellcolor{gray!15}56.3 \\
\arrayrulecolor{lightgray}\hhline{|~|--------------|} 
 & 512 & 1024 & \cellcolor{gray!15}SigLIP 2 & \cellcolor{gray!15}\bf{83.5} & \cellcolor{gray!15}\bf{77.8} & \cellcolor{gray!15}\bf{87.7} & \cellcolor{gray!15}\bf{84.6} & \cellcolor{gray!15}\bf{79.6} & \cellcolor{gray!15}\underline{55.2} & \cellcolor{gray!15}\bf{72.1} & \cellcolor{gray!15}\bf{85.3} & \cellcolor{gray!15}\bf{95.8} & \cellcolor{gray!15}\bf{47.4} & \cellcolor{gray!15}\bf{56.7} \\
\arrayrulecolor{black}\hhline{|---------------|} 
\multirow[c]{4}{*}{So/14} & \multirow[c]{2}{*}{224} & \multirow[c]{2}{*}{256} & SigLIP \cite{siglip} & 82.2 & 76.0 & 87.1 & 80.5 & 78.2 & 50.8 & 69.0 & 76.6 & 90.7 & 16.0 & 22.8 \\
 &  &  & \cellcolor{gray!15}SigLIP 2 & \cellcolor{gray!15}\underline{83.2} & \cellcolor{gray!15}\underline{77.7} & \cellcolor{gray!15}\underline{87.8} & \cellcolor{gray!15}\underline{84.6} & \cellcolor{gray!15}\underline{79.5} & \cellcolor{gray!15}\underline{55.1} & \cellcolor{gray!15}\underline{71.5} & \cellcolor{gray!15}\underline{84.3} & \cellcolor{gray!15}\underline{94.6} & \cellcolor{gray!15}\underline{47.9} & \cellcolor{gray!15}\underline{57.5} \\
\arrayrulecolor{lightgray}\hhline{|~|--------------|} 
 & \multirow[c]{2}{*}{384} & \multirow[c]{2}{*}{729} & SigLIP \cite{siglip} & 83.2 & 77.1 & 87.5 & 82.9 & 79.4 & 52.0 & 70.2 & 80.5 & 93.5 & 17.8 & 26.6 \\
 &  &  & \cellcolor{gray!15}SigLIP 2 & \cellcolor{gray!15}\bf{84.1} & \cellcolor{gray!15}\bf{78.7} & \cellcolor{gray!15}\bf{88.1} & \cellcolor{gray!15}\bf{86.0} & \cellcolor{gray!15}\bf{80.4} & \cellcolor{gray!15}\bf{55.8} & \cellcolor{gray!15}\bf{71.7} & \cellcolor{gray!15}\bf{85.7} & \cellcolor{gray!15}\bf{94.9} & \cellcolor{gray!15}\bf{48.4} & \cellcolor{gray!15}\bf{57.5} \\
\arrayrulecolor{black}\hhline{|---------------|} 
\multirow[c]{4}{*}{So/16} & \multirow[c]{2}{*}{256} & \multirow[c]{2}{*}{256} & mSigLIP \cite{siglip} & 80.8 & 74.1 & 86.1 & 79.5 & 77.1 & 49.4 & 68.6 & 80.0 & 92.1 & \bf{50.0} & \bf{62.8} \\
 &  &  & \cellcolor{gray!15}SigLIP 2 & \cellcolor{gray!15}83.4 & \cellcolor{gray!15}77.8 & \cellcolor{gray!15}87.7 & \cellcolor{gray!15}84.8 & \cellcolor{gray!15}79.7 & \cellcolor{gray!15}55.4 & \cellcolor{gray!15}\bf{71.5} & \cellcolor{gray!15}84.4 & \cellcolor{gray!15}94.2 & \cellcolor{gray!15}48.1 & \cellcolor{gray!15}57.5 \\
\arrayrulecolor{lightgray}\hhline{|~|--------------|} 
 & 384 & 576 & \cellcolor{gray!15}SigLIP 2 & \cellcolor{gray!15}\underline{84.1} & \cellcolor{gray!15}\underline{78.4} & \cellcolor{gray!15}\bf{88.1} & \cellcolor{gray!15}\underline{85.8} & \cellcolor{gray!15}\underline{80.4} & \cellcolor{gray!15}\bf{56.0} & \cellcolor{gray!15}71.2 & \cellcolor{gray!15}\underline{85.3} & \cellcolor{gray!15}\bf{95.9} & \cellcolor{gray!15}\underline{48.3} & \cellcolor{gray!15}57.5 \\
\arrayrulecolor{lightgray}\hhline{|~|--------------|} 
 & 512 & 1024 & \cellcolor{gray!15}SigLIP 2 & \cellcolor{gray!15}\bf{84.3} & \cellcolor{gray!15}\bf{79.1} & \cellcolor{gray!15}\underline{88.1} & \cellcolor{gray!15}\bf{86.2} & \cellcolor{gray!15}\bf{80.5} & \cellcolor{gray!15}\underline{56.0} & \cellcolor{gray!15}\underline{71.3} & \cellcolor{gray!15}\bf{85.5} & \cellcolor{gray!15}\underline{95.4} & \cellcolor{gray!15}48.3 & \cellcolor{gray!15}\underline{57.6} \\
\arrayrulecolor{black}\hhline{|---------------|} 
\multirow[c]{2}{*}{H/14} & \multirow[c]{2}{*}{224} & \multirow[c]{2}{*}{256} & MetaCLIP \cite{xu2024demystifying} & \underline{80.5} & \underline{74.1} & -- & \underline{76.5} & -- & \underline{57.5} & -- & \underline{85.0} & -- & -- & -- \\
 &  &  & DFN \cite{fang2024dfn} & \bf{83.4} & \bf{77.3} & -- & \bf{76.5} & -- & \bf{63.1} & -- & \bf{86.5} & -- & -- & -- \\
\arrayrulecolor{black}\hhline{|---------------|} 
\multirow[c]{2}{*}{g/16} & 256 & 256 & \cellcolor{gray!15}SigLIP 2 & \cellcolor{gray!15}\underline{84.5} & \cellcolor{gray!15}\underline{79.2} & \cellcolor{gray!15}\underline{88.3} & \cellcolor{gray!15}\underline{87.1} & \cellcolor{gray!15}\underline{82.1} & \cellcolor{gray!15}\underline{55.7} & \cellcolor{gray!15}\underline{72.5} & \cellcolor{gray!15}\underline{85.3} & \cellcolor{gray!15}\underline{95.3} & \cellcolor{gray!15}\underline{48.2} & \cellcolor{gray!15}\bf{58.2} \\
\arrayrulecolor{lightgray}\hhline{|~|--------------|} 
 & 384 & 576 & \cellcolor{gray!15}SigLIP 2 & \cellcolor{gray!15}\bf{85.0} & \cellcolor{gray!15}\bf{79.8} & \cellcolor{gray!15}\bf{88.5} & \cellcolor{gray!15}\bf{88.0} & \cellcolor{gray!15}\bf{82.5} & \cellcolor{gray!15}\bf{56.1} & \cellcolor{gray!15}\bf{72.8} & \cellcolor{gray!15}\bf{86.0} & \cellcolor{gray!15}\bf{95.4} & \cellcolor{gray!15}\bf{48.6} & \cellcolor{gray!15}\underline{57.9} \\
\arrayrulecolor{black}
\bottomrule
\end{tabular}

\caption{
Zero-shot classification, 10-shot (10s) classification (on the validation set), and retrieval performance (recall@1) of SigLIP~2 along with several baselines. SigLIP~2 outperforms the baselines---often by a large margin---despite being multilingual. Note that DFN~\cite{fang2024dfn} relies on a data filtering network fine-tuned on ImageNet, COCO, and Flickr.
}
\label{tab:zero_shot_main}
\end{table*}

\subsubsection{Fixed-resolution variant}

To obtain fixed-resolution checkpoints at multiple resolutions, we resume the checkpoints (with sequence length 256 and patch size 16) at 95\% of training, resize the positional embedding to the target sequences length (and in some cases the patch embedding from patch size 16 to 14 with the pseudoinverse (PI)-resize strategy from~\cite{beyer2023flexivit}), and resume the training at the target resolution with all losses. We opt for this approach as the common strategy of fine-tuning the final checkpoint with smaller learning rate and without weight decay~\cite{siglip} did not lead to good results across all sizes and resolutions.

\subsubsection{Variable aspect and resolution (NaFlex)} \label{sec:naflex}

\begin{figure*}[t]
    \centering
    \includegraphics[width=\linewidth]{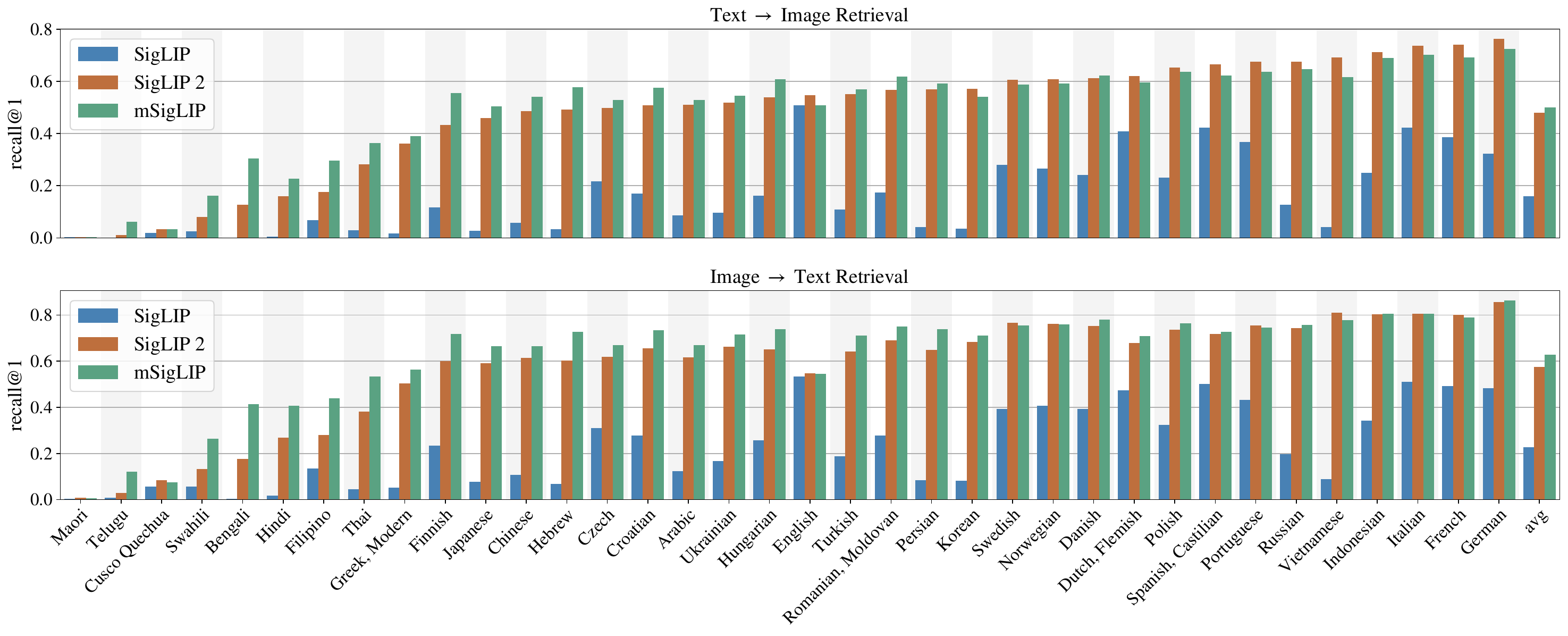}
    \caption{
    Per-language image-text retrieval performance for SigLIP, SigLIP~2 and mSigLIP on Crossmodal-3600~\cite{thapliyal-etal-2022-crossmodal-COCO35L-XM3600}. SigLIP~2 almost matches the performance of mSigLIP (SigLIP trained on multilingual data) despite performing substantially better on English vision-language tasks (Table~\ref{tab:zero_shot_main}).
    }
    \label{fig:xm3600}
\end{figure*}

NaFlex combines ideas from FlexiViT~\cite{beyer2023flexivit}, i.e. supporting multiple, predefined sequence lengths with a single ViT model, and NaViT~\cite{dehghani2024navit}, namely processing images at their native aspect ratio. This enables processing different types of images at appropriate resolution, e.g. using a larger resolution to process document images, while at the same time minimizing the impact of aspect ratio distortion on certain inference tasks, e.g. on OCR.

Given a patch size and target sequence length, NaFlex preprocesses the data by first resizing the input image such that the height and width after resizing are multiples of the patch size, while 1) keeping the aspect ratio distortion as small as possible and 2) producing a sequence length of at most the desired target sequence length. The resulting distortion in width and height is at most \texttt{(patch\_size-1)/width} and \texttt{(patch\_size-1)/height}, respectively, which tends to be small for common resolutions and aspect ratios. Note that NaViT incurs the same type of distortion. After resizing, the image is split into a sequence of patches, and patch coordinates as well as a mask with padding information is added (to handle the case where the actual sequence length is smaller than the target length).

To process different sequence lengths (and aspect ratios) with a ViT, we bilinearly resize (with anti-aliasing) the learned positional embedding to the target, non-square patch grid for the resized input image. We set the length of the learned positional embedding to 256, assuming a $16\times16$ patch grid before resizing. When the sequence length after resizing is smaller than the target sequence length, the attention layers (including the MAP head) are masked to ignore the extra padding tokens.

As for the fixed-resolution, adapted variants, we start from the default checkpoints trained with the setup described in Sec.~\ref{sec:siglip_training}, i.e. with non-aspect preserving resize to 256px, resulting in a sequence length of 256.
We take the checkpoint at 90\% training completion, then switch to aspect-preserving resizing and uniformly sampling a sequence length from $\{128, 256, 576, 784, 1024\}$ per mini-batch. At the same time we stretch the learning rate schedule corresponding to the last 10\% by a factor $3.75$ to ensure that each resolution is trained for sufficiently many examples. For the largest sequence length we further half the batch size and double the number of training steps to avoid out-of-memory errors. 

To keep implementation and computation complexity manageable, we do not apply self-distillation and masked prediction from Sec.~\ref{sec:tips}.

\begin{figure*}[t]
    \centering
    \includegraphics[width=\linewidth]{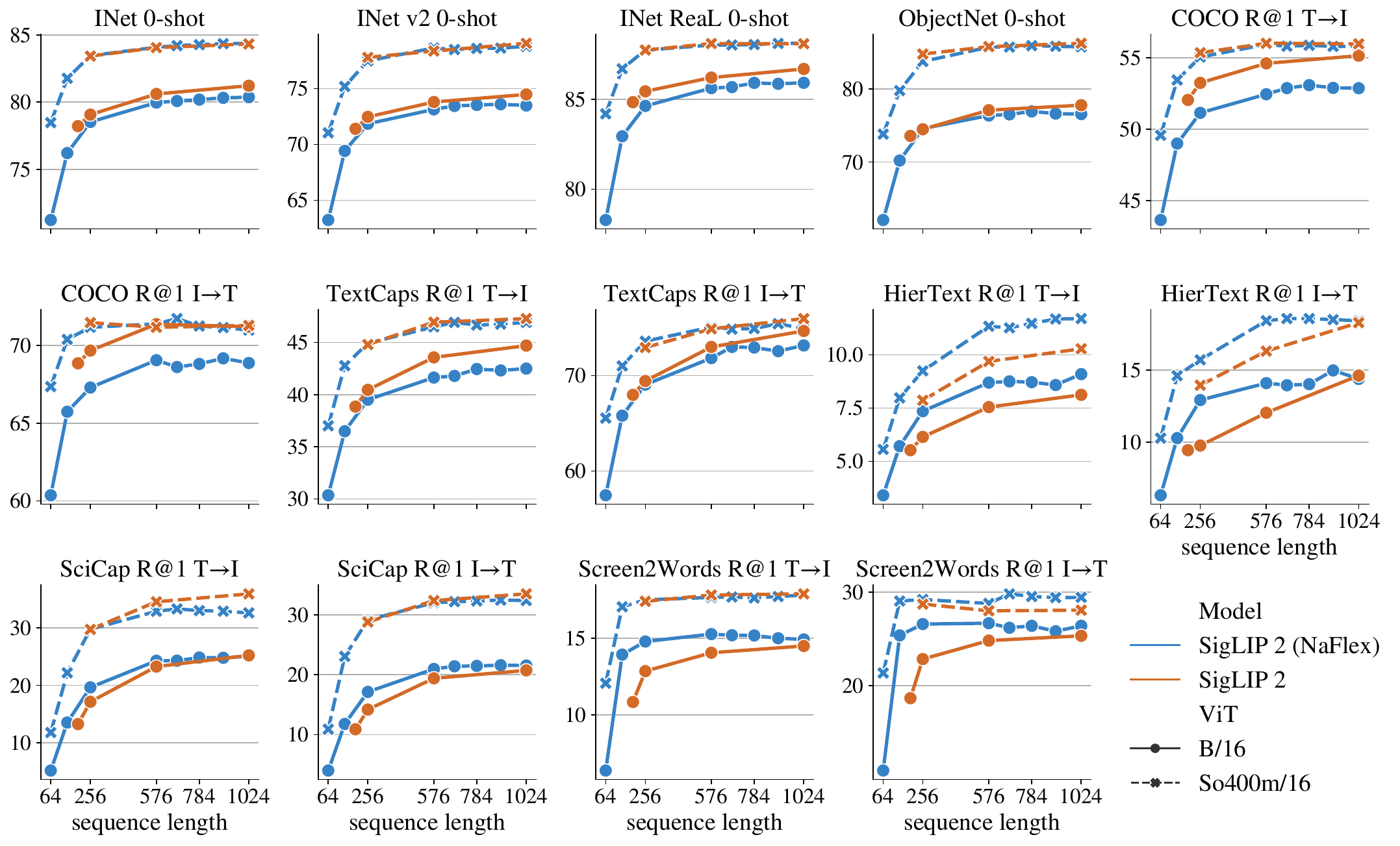}
    \caption{Comparing the NaFlex (a single checkpoint per model size supporting native aspect ratio and variable sequence length/resolution) and the standard square-input SigLIP~2 variants which use a separate checkpoint for each sequence length/resolution. The sequence lengths annotated on the x-axis correspond to training sequence lengths for NaFlex. NaFlex interpolates fairly well between training resolutions, but does not extrapolate well (not shown).}
    \label{fig:naflex}
\end{figure*}

\subsection{Distillation via active data curation} \label{sec:acid}

To maximize performance of the smallest fixed-resolution models (ViT-B/16 and ViT-B/32), we distill knowledge from a teacher (reference) model during a short fine-tuning stage. We lower the learning rate to $10^{-5}$, remove weight-decay, and continue training these models for an additional 4B examples using just the sigmoid image-text loss. During this stage, we perform implicit ``distillation through data'' using the ACID method proposed in~\cite{udandarao2024active}. Briefly, at every training step, the teacher model and the current learner model are used to score examples by their ``learnability''~\cite{mindermann2022prioritized}. These scores are then used to jointly select an optimal batch of size 32k from a larger super-batch~\cite{evansdata}. Here, we select data with a filtering ratio of 0.5 (i.e. super-batch size of 64k) to balance gains from curation with training compute. For the B/32 model, we find leveraging a filtering ratio of 0.75 is worth the extra cost.

We note that the authors in~\cite{udandarao2024active} suggest that the best performance is achieved with ACED, a method that combines ACID with explicit softmax-distillation (using a second teacher trained on more diverse data). However, here we propose a way to adapt ACID to capture these benefits \textit{without the need for explicit distillation}, saving significant amounts of compute. Specifically, instead of utilizing two separate teacher models, we take a single strong teacher trained on the diverse data (in this case, the SigLIP~2 So400m model) and fine-tune it for 1B examples on the high-quality curated dataset from~\cite{evansdata}. We then use this fine-tuned teacher model in the ACID method, as described above. Because this teacher blends diverse knowledge of concepts from pretraining, with knowledge of what is high-quality (from the curated dataset), the implicit distillation of ACID alone is sufficient to recover the benefits of ACED.

\begin{figure*}[t]
    \centering
    \includegraphics[width=\textwidth]{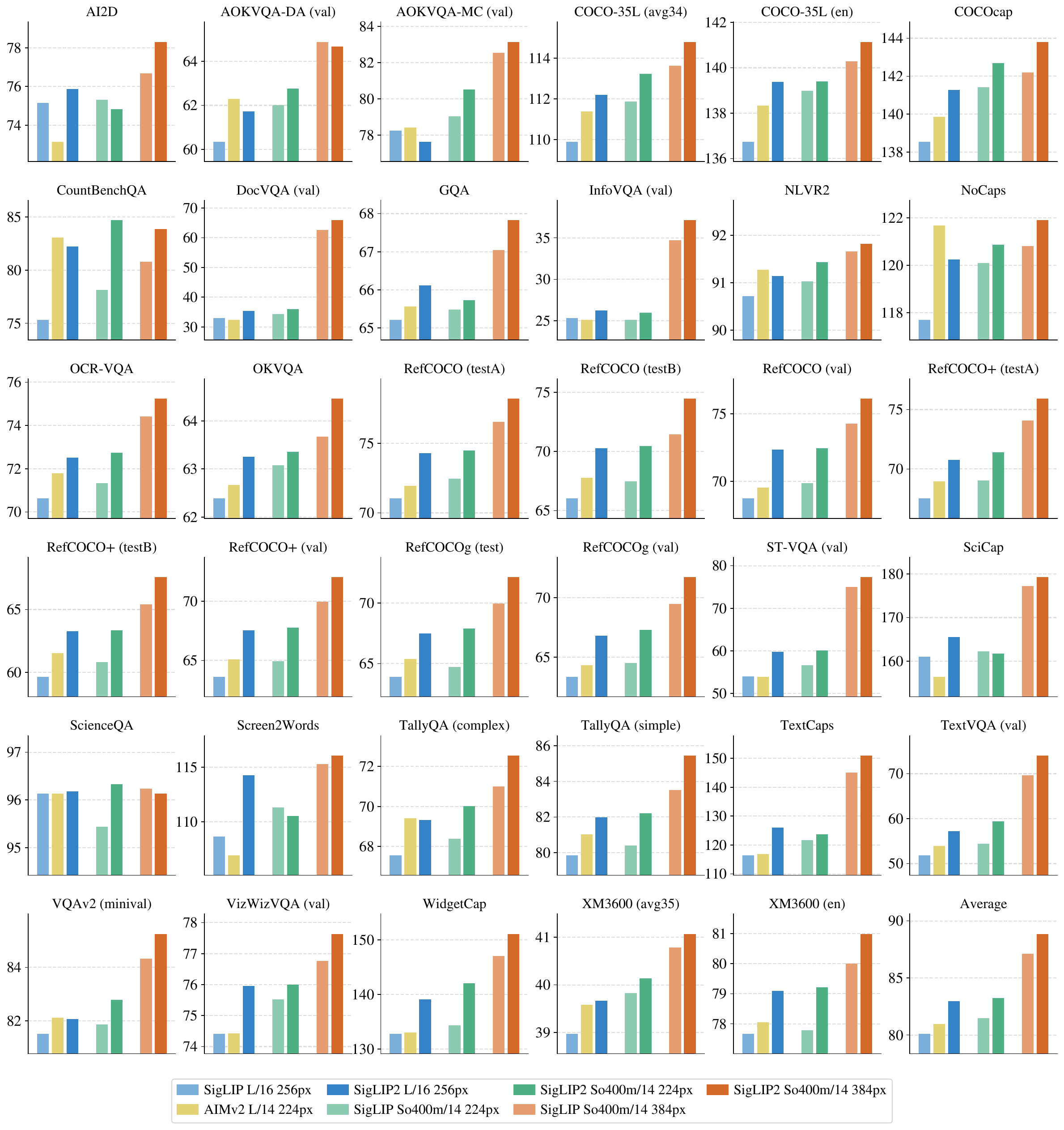}
    \caption{Comparison of different vision encoders after training a Gemma~2 LLM for 50M steps with a frozen vision encoder (PaliGemma~\cite{beyer2024paligemma} stage 1), followed by fine-tuning the VLM on individual datasets (PaliGemma stage 3). SigLIP~2 performs better than SigLIP and AIMv2~\cite{fini2024multimodal} for different model sizes and resolutions. Same data as in Table~\ref{tbl:paligemma}.}
    \label{fig:paligemma}
\end{figure*}

\section{Experiments and results} \label{sec:experiments}

\subsection{Zero-shot classification and retrieval}
In Table~\ref{tab:zero_shot_main} we report the performance of SigLIP~2 along with baselines on common zero-shot classification (ImageNet~\cite{deng2009imagenet} ObjectNet~\cite{barbu2019objectnet}, ImageNet-v2~\cite{recht2019imagenet}, ImageNet
ReaL~\cite{beyer2020we}) and image-text retrieval benchmarks. SigLIP~2 performs better than SigLIP and other (open-weight) baselines across the board, despite supporting many languages unlike the baselines (except mSigLIP~\cite{siglip}). Note that DFN~\cite{fang2024dfn}, which comes closest to SigLIP~2 on these benchmarks, uses a network fine-tuned on ImageNet, COCO, and Flickr (i.e. the main benchmarks in Table~\ref{tab:zero_shot_main}) as a filter to improve data quality. SigLIP~2's improvements over the baselines are particularly significant for the B-sized models owing to distillation (Sec.~\ref{sec:acid}). Moreover, we observe the common scaling trends as a function of image resolution and model size.

Table~\ref{tab:zero_shot_main} and Figure~\ref{fig:xm3600} further show the multilingual retrieval performance on Crossmodal-3600 (XM3600)~\cite{thapliyal-etal-2022-crossmodal-COCO35L-XM3600} covering 36 languages. SigLIP~2's recall exceeds that of SigLIP by a large margin, while only lagging slightly behind mSigLIP, which in turn performs substantially worse than SigLIP and SigLIP~2 on English-focused benchmarks.

\begin{table*}[t]
\footnotesize
\centering
\begin{tabular}{lccccccccc}
\toprule
 & & & \multicolumn{2}{c}{Segmentation $\uparrow$} & \multicolumn{2}{c}{Depth $\downarrow$} & \multicolumn{2}{c}{Normals $\downarrow$}\\
 \cmidrule(lr){4-5} \cmidrule(lr){6-7} \cmidrule(lr){8-9}
Model & ViT & Res. & PASCAL & ADE20k & NYUv2 & NAVI & NYUv2 & NAVI \\
\midrule
CLIP~\cite{clip} & L/14 & 224 & 74.5 & 39.0 & 0.553 & 0.073 & \bf{24.3} & 25.5 \\
OpenCLIP~\cite{ilharco2021open} & G/14 & 224 & 71.4 & 39.3 & 0.541 & -- & -- & -- \\
SigLIP~\cite{siglip} & So/14 & 224 & 72.0 & 37.6 & 0.576 & 0.083 & 25.9 & 26.0 \\
SigLIP 2 & So/14 & 224 & \bf{77.1} & \bf{41.8} & \bf{0.493} & \bf{0.067} & 24.9 & \bf{25.4} \\ \midrule
SigLIP~\cite{siglip} & So/14 & 384 & 73.8 & 40.8 & 0.563 & 0.069 & 24.1 & 25.4 \\
SigLIP 2 & So/14 & 384 & \bf{78.1} & \bf{45.4} & \bf{0.466} & \bf{0.064} & \bf{23.0} & \bf{25.0} \\
\bottomrule
\end{tabular}
\caption{Probing the frozen SigLIP~2 representation for a range of dense prediction tasks (metrics: segmentation: mIoU; depth: RMSE; normals; angular RMSE). SigLIP~2 outperforms several other popular open-weight models, often by a significant margin.}
\label{tab:dense_prediction}
\end{table*}

\subsubsection{NaFlex variant}
Fig.~\ref{fig:naflex} compares the fixed-resolution square-aspect ratio (standard) SigLIP~2 with the aspect-preserving NaFlex variant (one checkpoint for all sequence lengths) as a function of the sequence length. In addition to the retrieval benchmarks listed in the previous section, we add a range of OCR/document/screen-focused image-text benchmarks, namely TextCaps~\cite{sidorov2019textcaps}, HierText~\cite{long2023icdar}, SciCap~\cite{hsu2021scicap} and Screen2Words~\cite{wang2021screen2words}. The NaFlex variant outperforms the standard variant on the majority of these retrieval benchmarks, in particular for small sequence lengths (and hence resolutions) which tend to suffer more from aspect ratio distortion. On benchmarks predominantly based on natural images, the standard B-sized variant outperforms NaFlex, arguably thanks to the distillation step, whereas for the So400m architecture the two are on par. This is remarkable since the standard variant also benefits from the self-distillation stage (Sec.~\ref{sec:tips}).

\subsection{SigLIP~2 as a vision encoder for VLMs}
A popular use case for vision encoders like CLIP and SigLIP is to extract visual representations for VLMs~\cite{qwen-vl, blip2, peng2023kosmos, llava1, beyer2024paligemma, mm1, cambrian1}. The common paradigm combines a pretrained vision encoder with a pretrained LLM and does multimodal training on a rich mixture of vision language tasks. To evaluate the performance of SigLIP~2 in this application, we develop a recipe similar to that of PaliGemma 2 \cite{steiner2024paligemma}. Concretely, we combine SigLIP~2 vision encoders and baselines with the Gemma 2 2B LLM~\cite{gemmateam2024gemma2} and train the LLM on 50M examples of the Stage 1 training mix from~\cite{beyer2024paligemma, steiner2024paligemma} involving captioning, OCR, grounded captioning, visual question answering, detection, and instance segmentation (the annotations for the last 4 tasks are machine-generated, see~\cite[Sec.~3.2.5]{beyer2024paligemma} for details). We keep the vision encoder frozen (which has essentially no impact on quality \cite[Sec.~5.4]{beyer2024paligemma}) and reduce training duration to reflect a typical open model use case. The resulting VLM is then fine-tuned on a broad range of downstream tasks with the transfer settings from~\cite{steiner2024paligemma}. To understand the effect of the input resolution we perform experiments at resolution 224 or 256 (for models with patch size 14 and 16, respectively, to extract 256 image tokens) and 384px, but unlike~\cite{beyer2024paligemma, steiner2024paligemma} we repeat stage 1 at 384px rather than starting from the 224px variant.

Fig.~\ref{fig:paligemma} shows the results after fine-tuning for each dataset. Overall, SigLIP~2 clearly outperforms SigLIP across resolutions and model size. For an L-sized vision encoder, SigLIP~2 also outperforms the recently released AIMv2 model~\cite{fini2024multimodal}. The data from Fig.~\ref{fig:paligemma} can also be found in Table~\ref{tbl:paligemma}.

\begin{table*}[t]
\footnotesize
\centering
\begin{tabular}{lccccccc}
\toprule
Model & ViT & A-847 & PC-459 & A-150 & PC-59 & VOC-20 & VOC-21 \\
\midrule
CLIP~\cite{clip} & L/16 & 10.8 & 20.4 & 31.5 & 62.0 & 96.6 & 81.8 \\
OpenCLIP~\cite{ilharco2021open} & G/14 & 13.3 & 21.4 & 36.2 & 61.5 & \textbf{97.1} & 81.4 \\
SigLIP~\cite{siglip} & L/16 & 14.0 &	23.9 &	37.5 &	61.6 &	96.1 &	81.1 \\
SigLIP 2 & L/16 & \textbf{14.3} & \textbf{24.1} &	\textbf{38.8} & \textbf{62.4} &	97.0 & \textbf{82.3} \\
\bottomrule
\end{tabular}
\caption{We use Cat-Seg~\cite{catseg} to compare open-vocabulary segmentation performance (mIoU) of several models similar to 
\cite{naeem2024silc}. We observe that SigLIP~2 offers respectable improvements over comparable and even bigger models.
}
\label{tab:open_vocab_segmentation}
\end{table*}

\subsection{Dense prediction tasks}

\subsubsection{Semantic segmentation, depth estimation, surface normal estimation}

We adopt the evaluation protocol from~\cite{maninis2024tips} and probe the frozen SigLIP~2 representation, either with a linear layer or with a DPT decoder~\cite{ranftl2021vision}, on six benchmarks spanning semantic segmentation, monocular depth estimation, and surface normal estimation (see \citep[Sec.~4.1]{maninis2024tips} for details on the protocol and hyper parameters). Note, we make one (necessary) change: where the original method concatenates the CLS token to each of the patch feature vectors, we concatenate the output embedding of the MAP head instead, as we use a MAP head instead of a CLS token. The results in Table~\ref{tab:dense_prediction} indicate that SigLIP~2 outperforms several previous open, CLIP-style vision encoders, including SigLIP, often by a significant margin.

\subsubsection{Open-vocabulary segmentation}
Open-vocabulary segmentation aims to develop models that can segment any novel classes beyond a fixed training vocabulary. Here, we evaluate SigLIP~2's performance on this task. We use Cat-Seg~\cite{catseg} as a framework and compare performance across different models as proposed in ~\cite{naeem2024silc}. We train Cat-Seg on COCO-Stuff-164k~\cite{cocostuff} with 172 classes and then test it on various representative datasets with different vocabularies: ADE20k~\cite{ade20k1, ade20k2} with 847 or 150 classes (A-847/A-150), Pascal Context
(PC-459/PC-59)~\cite{context}, and Pascal VOC (VOC-20/VOC-21)~\cite{everingham2010pascal}.
The results can be found in Table~\ref{tab:open_vocab_segmentation}. We observe that the SigLIP~2 at L/16 improves on SigLIP and even surpasses the much bigger OpenCLIP G/14 model~\cite{ilharco2021open}.

\subsection{Localization tasks}

\subsubsection{Referring expression comprehension}

To probe the referring expression comprehension capabilities of SigLIP~2 on different RefCOCO variants~\cite{kazemzadeh2014referit, yu2016modeling} we apply the evaluation protocol from~\cite{locca}. We attach a 6-layer transformer decoder to the un-pooled, frozen vision encoder representation via cross-attention and train it from scratch on a mix of all RefCOCO variants (see~\cite{locca} for details). The results in Table~\ref{tab:refcocos} show that SigLIP~2 outperforms SigLIP as well as CLIP and pretraining via image captioning (Cap) by a large margin, across resolutions and model sizes. This can be attributed to the decoder-based pretraining, as described in Sec.~\ref{sec:siglip_training}. SigLIP~2 is only outperformed LocCa, which we hypothesize might be due to the fact that SigLIP~2 is pretrained on multilingual data. LocCa, on the other hand, is trained on text only from English web sites. Finally, note that we expect significant improvements when using the decoder from pretraining as observed for LocCa.

\subsubsection{Open-vocabulary detection}

OWL-ViT~\cite{minderer2022simple} is a popular method to adapt CLIP-style vision-language models to open-vocabulary detection. Here, we apply this approach to SigLIP and SigLIP~2 models, closely following the data and optimizer configuration from~\cite{minderer2022simple}. The results in Table~\ref{tab:owlvit} show that SigLIP~2 achieves better performance than SigLIP on the two popular benchmarks COCO~\cite{coco2014} and LVIS~\cite{gupta2019lvis}. The relative improvement is most pronounced for the LVIS rare categories. Further, the results here are better than those in~\cite{minderer2022simple} which is likely because \cite{minderer2022simple} used CLIP rather than SigLIP.

\subsection{Cultural diversity and fairness}

Besides the improvement in model quality in SigLIP~2 compared to its predecessor, SigLIP~2 is also more inclusive in two aspects. First, we follow the recommendations of~\cite{pouget2024no} and utilize a training mixture comprising both English and multilingual data to enhance cultural diversity. Second, to address potential societal biases in the training data, we integrate the data de-biasing techniques from~\cite{alabdulmohsin2024clip}. These techniques are applied to mitigate biases in both first-order statistics, such as disparities in gender representation, and second-order statistics, such as biased associations between gender and occupation. Next, we present the evaluation results.

\begin{table}[t]
\footnotesize
\setlength{\tabcolsep}{0.5em}
\begin{tabular}{llccc}
\toprule
 ViT & Model & COCO (AP) & LVIS (AP) & LVIS (APr) \\
\midrule
\multirow[c]{2}{*}{B/16} & SigLIP & 42.2 & 33.0 & 31.0 \\
 & SigLIP 2 & \bf{42.8} & \bf{34.4} & \bf{32.7} \\
\midrule
\multirow[c]{2}{*}{So/14} & SigLIP & 44.3 & 39.5 & 40.9 \\
 & SigLIP 2 & \bf{45.2} & \bf{40.5} & \bf{42.3} \\
\bottomrule
\end{tabular}
\caption{Fine-tuned SigLIP and SigLIP~2 for open-vocabulary detection via OWL-ViT~\cite{minderer2022simple}.}
\label{tab:owlvit}
\end{table}

\begin{table*}[t]
\vspace{-0.2cm}
\footnotesize
\centering
\begin{tabular}{lclcccccccc}
\toprule
 &  &  & \multicolumn{3}{c}{RefCOCO} & \multicolumn{3}{c}{RefCOCO+} & \multicolumn{2}{c}{RefCOCOg} \\
\cmidrule(lr){4-6} \cmidrule(lr){7-9} \cmidrule(lr){10-11}
 ViT &  Seq. & Model & val & testA & testB & val & testA & testB & val-u & test-u \\
\midrule
\multirow[c]{4}{*}{B} & \multirow[c]{2}{*}{256} & SigLIP \cite{siglip} & 64.05 & 70.10 & 57.89 & 55.77 & 63.57 & 47.51 & 59.06 & 60.33 \\
 &  & \cellcolor{gray!15}SigLIP 2 & \cellcolor{gray!15}83.76 & \cellcolor{gray!15}86.21 & \cellcolor{gray!15}79.57 & \cellcolor{gray!15}74.26 & \cellcolor{gray!15}79.85 & \cellcolor{gray!15}65.83 & \cellcolor{gray!15}77.25 & \cellcolor{gray!15}77.83 \\
\arrayrulecolor{lightgray}\hhline{|~|----------|}
 & \multirow[c]{2}{*}{576} & SigLIP \cite{siglip} & 67.17 & 72.94 & 60.94 & 59.09 & 67.26 & 50.22 & 61.98 & 62.64 \\
 &  & \cellcolor{gray!15}SigLIP 2 & \cellcolor{gray!15}\bf{85.18} & \cellcolor{gray!15}\bf{87.92} & \cellcolor{gray!15}\bf{80.53} & \cellcolor{gray!15}\bf{76.08} & \cellcolor{gray!15}\bf{82.17} & \cellcolor{gray!15}\bf{67.10} & \cellcolor{gray!15}\bf{79.08} & \cellcolor{gray!15}\bf{79.60} \\
\arrayrulecolor{black}\hhline{|-----------|} 
\multirow[c]{8}{*}{L} & \multirow[c]{6}{*}{256} & Cap \cite{cappa} & 60.64 & 65.47 & 56.17 & 52.56 & 58.32 & 45.99 & 56.75 & 57.99 \\
 &  & CapPa \cite{cappa} & 64.17 & 69.90 & 58.25 & 56.14 & 63.68 & 48.18 & 58.90 & 59.91 \\
 &  & CLIP \cite{clip} & 65.21 & 71.28 & 58.17 & 57.53 & 66.44 & 47.77 & 59.32 & 60.24 \\
 &  & SigLIP \cite{siglip} & 67.33 & 72.40 & 61.21 & 59.57 & 67.09 & 51.08 & 61.89 & 62.90 \\
 &  & \cellcolor{gray!15}SigLIP 2 & \cellcolor{gray!15}86.04 & \cellcolor{gray!15}89.02 & \cellcolor{gray!15}81.85 & \cellcolor{gray!15}77.29 & \cellcolor{gray!15}83.28 & \cellcolor{gray!15}70.16 & \cellcolor{gray!15}80.11 & \cellcolor{gray!15}80.78 \\
 &  & LocCa \cite{locca} & \bf{88.34} & \bf{91.20} & \bf{85.10} & \bf{79.39} & \bf{85.13} & \bf{72.61} & \underline{81.69} & \bf{82.64} \\
\arrayrulecolor{lightgray}\hhline{|~|----------|}
 & \multirow[c]{2}{*}{576} & SigLIP \cite{siglip} & 70.76 & 76.32 & 63.79 & 63.38 & 71.48 & 54.65 & 64.73 & 65.74 \\
 &  & \cellcolor{gray!15}SigLIP 2 & \cellcolor{gray!15}\underline{87.28} & \cellcolor{gray!15}\underline{90.29} & \cellcolor{gray!15}\underline{82.85} & \cellcolor{gray!15}\underline{79.00} & \cellcolor{gray!15}\underline{85.00} & \cellcolor{gray!15}\underline{70.92} & \cellcolor{gray!15}\bf{81.84} & \cellcolor{gray!15}\underline{82.15} \\
\arrayrulecolor{black}\hhline{|-----------|} 
\multirow[c]{4}{*}{So} & \multirow[c]{2}{*}{256} & SigLIP \cite{siglip} & 64.68 & 71.23 & 58.40 & 57.43 & 66.06 & 49.38 & 59.66 & 60.88 \\
 &  & \cellcolor{gray!15}SigLIP 2 & \cellcolor{gray!15}86.42 & \cellcolor{gray!15}89.41 & \cellcolor{gray!15}82.48 & \cellcolor{gray!15}77.81 & \cellcolor{gray!15}84.36 & \cellcolor{gray!15}70.67 & \cellcolor{gray!15}80.83 & \cellcolor{gray!15}81.27 \\
\arrayrulecolor{lightgray}\hhline{|~|----------|}
 & \multirow[c]{2}{*}{729} & SigLIP \cite{siglip} & 67.66 & 74.12 & 62.36 & 60.74 & 69.73 & 52.12 & 62.61 & 63.24 \\
 &  & \cellcolor{gray!15}SigLIP 2 & \cellcolor{gray!15}\bf{87.88} & \cellcolor{gray!15}\bf{91.13} & \cellcolor{gray!15}\bf{83.59} & \cellcolor{gray!15}\bf{80.06} & \cellcolor{gray!15}\bf{86.30} & \cellcolor{gray!15}\bf{72.66} & \cellcolor{gray!15}\bf{82.68} & \cellcolor{gray!15}\bf{83.63} \\
\arrayrulecolor{black}\hhline{|-----------|} 
\multirow[c]{2}{*}{g} & 256 & \cellcolor{gray!15}SigLIP 2 & \cellcolor{gray!15}87.31 & \cellcolor{gray!15}90.24 & \cellcolor{gray!15}83.25 & \cellcolor{gray!15}79.25 & \cellcolor{gray!15}85.23 & \cellcolor{gray!15}71.60 & \cellcolor{gray!15}81.48 & \cellcolor{gray!15}82.14 \\
\arrayrulecolor{lightgray}\hhline{|~|----------|}
 & 576 & \cellcolor{gray!15}SigLIP 2 & \cellcolor{gray!15}\bf{88.45} & \cellcolor{gray!15}\bf{91.53} & \cellcolor{gray!15}\bf{84.95} & \cellcolor{gray!15}\bf{80.44} & \cellcolor{gray!15}\bf{87.09} & \cellcolor{gray!15}\bf{73.53} & \cellcolor{gray!15}\bf{83.12} & \cellcolor{gray!15}\bf{84.14} \\
\arrayrulecolor{black} 
\bottomrule
\end{tabular}

\caption{
Comparing SigLIP~2 models with SigLIP and other baselines from the literature on referring expression comprehension (Acc@0.5). For matching model size and sequence length (seq.) SigLIP~2 models outperform SigLIP models substantially. SigLIP~2 is only outperformed by LocCa, which uses the same decoder-based loss, but is trained on captions from English language websites only.
}\vspace{-0.2cm}
\label{tab:refcocos}
\end{table*}

\paragraph{Cultural Diversity} To evaluate for cultural diversity, we report the zero-shot classification accuracy results using Dollar Street~\cite{rojas2022dollar}, GeoDE~\cite{ramaswamy2024geode}, and Google Landmarks Dataset v2 (GLDv2)~\cite{weyand2020google}. We also include 10-shot geolocalization using Dollar Street and GeoDE, as proposed in~\cite{pouget2024no}. For zero-shot evaluation on Dollar Street, we implement the methodology outlined in~\cite{rojas2022dollar}, mapping 96 topics within the dataset to corresponding ImageNet classes. This process results in a subset of 21K images for our analysis.

Fig.~\ref{fig:cultural_diversity} shows a set of representative results (full results are shown in Appendix~\ref{app:cultural_diversity}). We observe an improvement in these metrics in SigLIP~2 compared to SigLIP for the same model size and resolution, and the improvements are particularly significant in geolocalization tasks. For instance, 10-shot geolocalization accuracy in GeoDE (region) improves from 36.2\% for SigLIP L/16 at 256px to 44.4\% in SigLIP~2. Similarly, 0-shot accuracy on Dollar Street improves from 52.1\% to 55.2\% in the same models.

\begin{figure}[t]
    \centering
    \includegraphics[width=0.8\columnwidth]{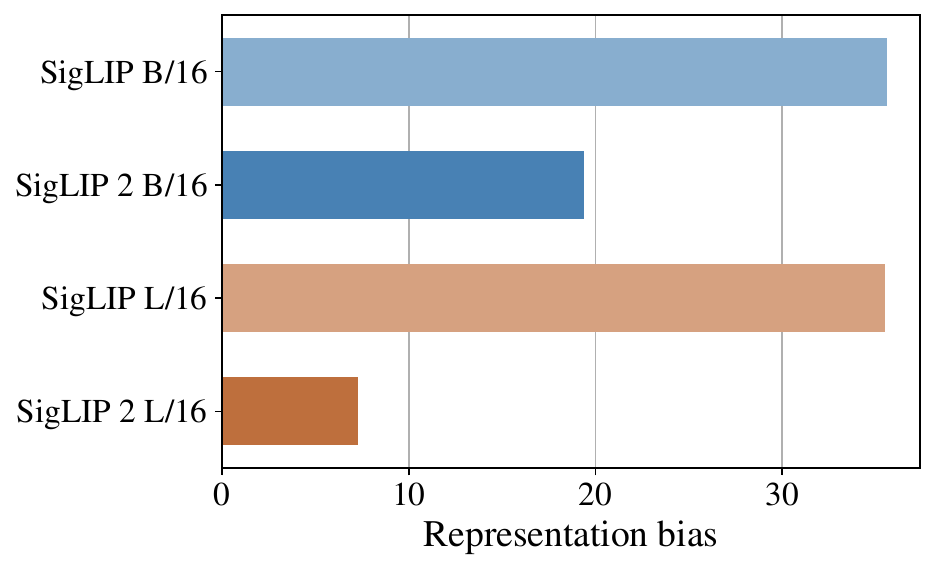}
    \caption{Representation bias (association of random objects with gender; lower is better) for different models.}
    \label{fig:rep_bias}
\end{figure}

\paragraph{Fairness} In terms of fairness, we report two metrics. The first is ``representation bias,'' as defined in~\cite{alabdulmohsin2024clip}, which measures the tendency in the model to associate a random object (such as cars) with a particular gender group. As shown in Fig.~\ref{fig:rep_bias}, SigLIP~2 is \emph{significantly} better than SigLIP. For instance, while SigLIP L/16 at 256px has a representation bias of about 35.5\%---meaning it prefers to associate random images with ``men'' over ``women'' more than 85.5\% of the time---SigLIP~2 of the same size and resolution has a representation bias of  7.3\% only. In addition, larger models tend to exhibit less representation bias than smaller models, in agreement with the earlier findings in~\cite{alabdulmohsin2024clip}.

We also investigate the Dollar Street 0-shot results by income level and the GeoDE results by geographic region as \cite{pouget2024no}. However, in this context we only observe very minor benefits, or no benefits when comparing SigLIP and SigLIP~2 models of matching size and resolution (some results shown in Table~\ref{tbl:app_rb}).

\section{Related work}\label{sec:related}

Contrastive pretraining as popularized by CLIP~\cite{clip} and ALIGN~\cite{align} has become the dominant approach for learning high-level, semantic, visual representations that perform well on classification and retrieval, as vision encoders for VLMs~\cite{qwen-vl, blip2, peng2023kosmos, llava1, beyer2024paligemma, mm1, cambrian1} and open-vocabulary tasks including detection~\cite{minderer2022simple, kuo2023open, owlvitv2} and segmentation~\cite{ding2022decoupling, catseg}. Besides the original CLIP release, several projects have released open-weight contrastive models~\cite{ilharco2021open, sun2023eva, siglip, li2023clipa, fang2024dfn, xu2024demystifying}. At a high level, these works follow training methods that are relatively close to the original CLIP method, mainly \cite{siglip} proposing modified loss functions and \cite{fang2024dfn, xu2024demystifying} targeting data quality and filtering.

More generally, a large number of modifications and improvements  to contrastive training have been proposed in the literature. \cite{gadre2024datacomp, fang2024dfn, xu2024demystifying, evansdata, udandarao2024active} study filtering techniques to improve data quality. With a similar motivation, \cite{fan2023improving, nguyen2024improving, lai2024veclip, maninis2024tips} re-caption training images with VLMs to improve the caption quality and hence the quality of the training signal. Another promising area has been to modify or augment the loss function. \cite{mu2022slip, naeem2024silc, maninis2024tips} combine CLIP with self-supervised losses. Another popular approach is to add a language decoder to train with captioning as an auxiliary task~\cite{yu2022coca, blip2}. Captioning as a standalone representation learning task has attracted less attention, but can produce visual representations competitive with contrastive training~\cite{wang2021simvlm, cappa, locca, fini2024multimodal}.

\begin{figure}[t]
    \centering
    \includegraphics[width=\columnwidth]{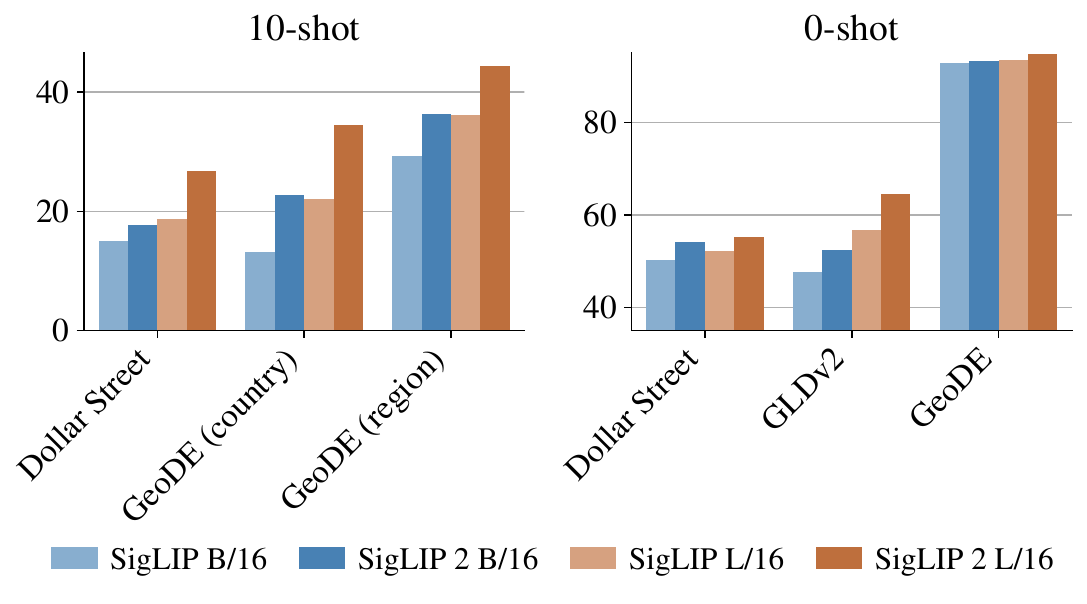}
    \caption{10-shot and 0-shot accuracy for geographically diverse object classification tasks (Dollar Street, GeoDE), as well as geolocalization (GeoDE country/region) and landmark localization (GLDv2) tasks. SigLIP~2 consistently performs better than SigLIP (see Table~\ref{tbl:app_cultural_diversity} for additional results).}
    \label{fig:cultural_diversity}
\end{figure}

\section{Conclusion}\label{sec:conclusion}
In this work, we introduced SigLIP~2, a family of open-weight multilingual vision-language encoders that builds on the success of SigLIP. By incorporating a combination of techniques such as decoder-based pretraining, self-supervised losses, and active data curation, SigLIP~2 achieves significant improvements in zero-shot classification, transfer performance as a vision encoder in VLMs, and in localization and dense prediction tasks. Furthermore, thanks to training on multilingual data and applying de-biasing filters, SigLIP~2 attains more balanced quality across culturally diverse data. Finally, the NaFlex variant enables the model to support multiple resolutions with a single model checkpoint, while preserving the native image aspect ratio. We hope that our SigLIP~2 release will enable many exciting applications within the open-source community.

\paragraph{Acknowledgments} We would like to thank Josip Djolonga, Neil Houlsby, Andre Araujo, Kevis Maninis, and Phoebe Kirk for discussions and feedback on this project. We also thank Joan Puigcerver, Andr\'e Susano Pinto, and Alex Bewley for infrastructure contributions to the \texttt{big\_vision} code base, which were helpful for this project.

\FloatBarrier

\bibliographystyle{abbrvnat}
\nobibliography*
\bibliography{document}

\clearpage

\appendix
\onecolumn
\section*{Appendix}

\section{Full PaliGemma results}
\vspace*{\fill}
\begin{table*}[h]
  \centering
  \footnotesize
  \begin{tabular}{lccccccc}
\toprule
 & \multicolumn{3}{c}{Large 224/256px} & \multicolumn{2}{c}{So400m/14 224px} & \multicolumn{2}{c}{So400m 384px} \\
\cmidrule(lr){2-4} \cmidrule(lr){5-6} \cmidrule(lr){7-8}
\rule{0pt}{10pt}
 & SigLIP & AIMv2 & SigLIP2 & SigLIP & SigLIP2 & SigLIP & SigLIP2 \\
\midrule

AI2D & \phantom{-}\phantom{0}75.2 & \phantom{-}\phantom{0}73.2 & \phantom{-}\phantom{0}75.9 & \phantom{-}\phantom{0}75.3 & \phantom{-}\phantom{0}74.8 & \phantom{-}\phantom{0}76.7 & \phantom{-}\phantom{0}78.3 \\
AOKVQA-DA (val) & \phantom{-}\phantom{0}60.3 & \phantom{-}\phantom{0}62.3 & \phantom{-}\phantom{0}61.7 & \phantom{-}\phantom{0}62.0 & \phantom{-}\phantom{0}62.8 & \phantom{-}\phantom{0}64.9 & \phantom{-}\phantom{0}64.7 \\
AOKVQA-MC (val) & \phantom{-}\phantom{0}78.3 & \phantom{-}\phantom{0}78.4 & \phantom{-}\phantom{0}77.6 & \phantom{-}\phantom{0}79.0 & \phantom{-}\phantom{0}80.5 & \phantom{-}\phantom{0}82.5 & \phantom{-}\phantom{0}83.1 \\
COCO-35L (avg34) & \phantom{-}109.9 & \phantom{-}111.4 & \phantom{-}112.2 & \phantom{-}111.9 & \phantom{-}113.2 & \phantom{-}113.6 & \phantom{-}114.8 \\
COCO-35L (en) & \phantom{-}136.7 & \phantom{-}138.3 & \phantom{-}139.4 & \phantom{-}139.0 & \phantom{-}139.4 & \phantom{-}140.3 & \phantom{-}141.1 \\
COCOcap & \phantom{-}138.6 & \phantom{-}139.9 & \phantom{-}141.3 & \phantom{-}141.4 & \phantom{-}142.7 & \phantom{-}142.2 & \phantom{-}143.8 \\
CountBenchQA & \phantom{-}\phantom{0}75.3 & \phantom{-}\phantom{0}83.1 & \phantom{-}\phantom{0}82.2 & \phantom{-}\phantom{0}78.2 & \phantom{-}\phantom{0}84.7 & \phantom{-}\phantom{0}80.8 & \phantom{-}\phantom{0}83.9 \\
DocVQA (val) & \phantom{-}\phantom{0}33.0 & \phantom{-}\phantom{0}32.3 & \phantom{-}\phantom{0}35.4 & \phantom{-}\phantom{0}34.3 & \phantom{-}\phantom{0}35.9 & \phantom{-}\phantom{0}62.7 & \phantom{-}\phantom{0}65.9 \\
GQA & \phantom{-}\phantom{0}65.2 & \phantom{-}\phantom{0}65.6 & \phantom{-}\phantom{0}66.1 & \phantom{-}\phantom{0}65.5 & \phantom{-}\phantom{0}65.7 & \phantom{-}\phantom{0}67.0 & \phantom{-}\phantom{0}67.8 \\
InfoVQA (val) & \phantom{-}\phantom{0}25.3 & \phantom{-}\phantom{0}25.1 & \phantom{-}\phantom{0}26.3 & \phantom{-}\phantom{0}25.1 & \phantom{-}\phantom{0}26.0 & \phantom{-}\phantom{0}34.7 & \phantom{-}\phantom{0}37.1 \\
NLVR2 & \phantom{-}\phantom{0}90.7 & \phantom{-}\phantom{0}91.3 & \phantom{-}\phantom{0}91.1 & \phantom{-}\phantom{0}91.0 & \phantom{-}\phantom{0}91.4 & \phantom{-}\phantom{0}91.7 & \phantom{-}\phantom{0}91.8 \\
NoCaps & \phantom{-}117.7 & \phantom{-}121.7 & \phantom{-}120.3 & \phantom{-}120.1 & \phantom{-}120.9 & \phantom{-}120.8 & \phantom{-}121.9 \\
OCR-VQA & \phantom{-}\phantom{0}70.6 & \phantom{-}\phantom{0}71.8 & \phantom{-}\phantom{0}72.5 & \phantom{-}\phantom{0}71.3 & \phantom{-}\phantom{0}72.7 & \phantom{-}\phantom{0}74.4 & \phantom{-}\phantom{0}75.2 \\
OKVQA & \phantom{-}\phantom{0}62.4 & \phantom{-}\phantom{0}62.7 & \phantom{-}\phantom{0}63.3 & \phantom{-}\phantom{0}63.1 & \phantom{-}\phantom{0}63.4 & \phantom{-}\phantom{0}63.7 & \phantom{-}\phantom{0}64.5 \\
RefCOCO (testA) & \phantom{-}\phantom{0}71.0 & \phantom{-}\phantom{0}71.9 & \phantom{-}\phantom{0}74.3 & \phantom{-}\phantom{0}72.4 & \phantom{-}\phantom{0}74.5 & \phantom{-}\phantom{0}76.6 & \phantom{-}\phantom{0}78.2 \\
RefCOCO (testB) & \phantom{-}\phantom{0}66.0 & \phantom{-}\phantom{0}67.8 & \phantom{-}\phantom{0}70.3 & \phantom{-}\phantom{0}67.5 & \phantom{-}\phantom{0}70.5 & \phantom{-}\phantom{0}71.4 & \phantom{-}\phantom{0}74.5 \\
RefCOCO (val) & \phantom{-}\phantom{0}68.7 & \phantom{-}\phantom{0}69.5 & \phantom{-}\phantom{0}72.4 & \phantom{-}\phantom{0}69.9 & \phantom{-}\phantom{0}72.5 & \phantom{-}\phantom{0}74.3 & \phantom{-}\phantom{0}76.1 \\
RefCOCO+ (testA) & \phantom{-}\phantom{0}67.5 & \phantom{-}\phantom{0}69.0 & \phantom{-}\phantom{0}70.8 & \phantom{-}\phantom{0}69.0 & \phantom{-}\phantom{0}71.4 & \phantom{-}\phantom{0}74.1 & \phantom{-}\phantom{0}75.9 \\
RefCOCO+ (testB) & \phantom{-}\phantom{0}59.6 & \phantom{-}\phantom{0}61.5 & \phantom{-}\phantom{0}63.3 & \phantom{-}\phantom{0}60.8 & \phantom{-}\phantom{0}63.3 & \phantom{-}\phantom{0}65.4 & \phantom{-}\phantom{0}67.6 \\
RefCOCO+ (val) & \phantom{-}\phantom{0}63.6 & \phantom{-}\phantom{0}65.1 & \phantom{-}\phantom{0}67.6 & \phantom{-}\phantom{0}64.9 & \phantom{-}\phantom{0}67.8 & \phantom{-}\phantom{0}70.0 & \phantom{-}\phantom{0}72.0 \\
RefCOCOg (test) & \phantom{-}\phantom{0}63.9 & \phantom{-}\phantom{0}65.4 & \phantom{-}\phantom{0}67.5 & \phantom{-}\phantom{0}64.7 & \phantom{-}\phantom{0}67.9 & \phantom{-}\phantom{0}69.9 & \phantom{-}\phantom{0}72.1 \\
RefCOCOg (val) & \phantom{-}\phantom{0}63.3 & \phantom{-}\phantom{0}64.3 & \phantom{-}\phantom{0}66.8 & \phantom{-}\phantom{0}64.5 & \phantom{-}\phantom{0}67.3 & \phantom{-}\phantom{0}69.5 & \phantom{-}\phantom{0}71.7 \\
ST-VQA (val) & \phantom{-}\phantom{0}54.0 & \phantom{-}\phantom{0}53.9 & \phantom{-}\phantom{0}59.8 & \phantom{-}\phantom{0}56.7 & \phantom{-}\phantom{0}60.1 & \phantom{-}\phantom{0}75.0 & \phantom{-}\phantom{0}77.3 \\
SciCap & \phantom{-}161.1 & \phantom{-}156.4 & \phantom{-}165.5 & \phantom{-}162.3 & \phantom{-}161.8 & \phantom{-}177.2 & \phantom{-}179.3 \\
ScienceQA & \phantom{-}\phantom{0}96.1 & \phantom{-}\phantom{0}96.1 & \phantom{-}\phantom{0}96.2 & \phantom{-}\phantom{0}95.4 & \phantom{-}\phantom{0}96.3 & \phantom{-}\phantom{0}96.2 & \phantom{-}\phantom{0}96.1 \\
Screen2Words & \phantom{-}108.7 & \phantom{-}106.9 & \phantom{-}114.3 & \phantom{-}111.3 & \phantom{-}110.6 & \phantom{-}115.3 & \phantom{-}116.1 \\
TallyQA (complex) & \phantom{-}\phantom{0}67.6 & \phantom{-}\phantom{0}69.4 & \phantom{-}\phantom{0}69.3 & \phantom{-}\phantom{0}68.4 & \phantom{-}\phantom{0}70.0 & \phantom{-}\phantom{0}71.0 & \phantom{-}\phantom{0}72.5 \\
TallyQA (simple) & \phantom{-}\phantom{0}79.9 & \phantom{-}\phantom{0}81.0 & \phantom{-}\phantom{0}82.0 & \phantom{-}\phantom{0}80.4 & \phantom{-}\phantom{0}82.2 & \phantom{-}\phantom{0}83.5 & \phantom{-}\phantom{0}85.4 \\
TextCaps & \phantom{-}116.5 & \phantom{-}116.8 & \phantom{-}126.1 & \phantom{-}121.7 & \phantom{-}123.8 & \phantom{-}145.0 & \phantom{-}150.9 \\
TextVQA (val) & \phantom{-}\phantom{0}51.9 & \phantom{-}\phantom{0}53.9 & \phantom{-}\phantom{0}57.3 & \phantom{-}\phantom{0}54.5 & \phantom{-}\phantom{0}59.4 & \phantom{-}\phantom{0}69.7 & \phantom{-}\phantom{0}74.0 \\
VQAv2 (minival) & \phantom{-}\phantom{0}81.5 & \phantom{-}\phantom{0}82.1 & \phantom{-}\phantom{0}82.1 & \phantom{-}\phantom{0}81.9 & \phantom{-}\phantom{0}82.8 & \phantom{-}\phantom{0}84.3 & \phantom{-}\phantom{0}85.2 \\
VizWizVQA (val) & \phantom{-}\phantom{0}74.4 & \phantom{-}\phantom{0}74.4 & \phantom{-}\phantom{0}76.0 & \phantom{-}\phantom{0}75.5 & \phantom{-}\phantom{0}76.0 & \phantom{-}\phantom{0}76.8 & \phantom{-}\phantom{0}77.6 \\
WidgetCap & \phantom{-}132.8 & \phantom{-}133.0 & \phantom{-}139.1 & \phantom{-}134.4 & \phantom{-}142.0 & \phantom{-}147.0 & \phantom{-}151.1 \\
XM3600 (avg35) & \phantom{-}\phantom{0}39.0 & \phantom{-}\phantom{0}39.6 & \phantom{-}\phantom{0}39.7 & \phantom{-}\phantom{0}39.8 & \phantom{-}\phantom{0}40.1 & \phantom{-}\phantom{0}40.8 & \phantom{-}\phantom{0}41.1 \\
XM3600 (en) & \phantom{-}\phantom{0}77.7 & \phantom{-}\phantom{0}78.0 & \phantom{-}\phantom{0}79.1 & \phantom{-}\phantom{0}77.8 & \phantom{-}\phantom{0}79.2 & \phantom{-}\phantom{0}80.0 & \phantom{-}\phantom{0}81.0 \\
\bottomrule
\end{tabular}

  \caption{The first three columns compare Large-sized models with 256 tokens each (that's 224px for the AIMv2 model with patch size 14, and 256px for the SigLIP models with patch size 16). The last four columns compare So400M-sized SigLIP models with patch size 14 at two different resolutions (and hence tokens). Same data as in Figure~\ref{fig:paligemma}.}
  \label{tbl:paligemma}
\end{table*}
\vspace*{\fill}

\clearpage

\section{Full NaFlex results}
\vspace*{\fill}
\begin{table*}[h]
  \centering
  \footnotesize
  \setlength{\tabcolsep}{0.35em}
\begin{tabular}{lrlcccccccccccccc}
\toprule
 &  &  & \multicolumn{4}{c}{ImageNet-1k} & \multicolumn{2}{c}{COCO R@1} & \multicolumn{2}{c}{TC R@1} & \multicolumn{2}{c}{HT R@1} & \multicolumn{2}{c}{SC R@1} & \multicolumn{2}{c}{S2W R@1} \\
 \cmidrule(lr){4-7}\cmidrule(lr){8-9}\cmidrule(lr){10-11}\cmidrule(lr){12-13}\cmidrule(lr){14-15}\cmidrule(lr){16-17}
ViT & Seq. & Model & val & v2 & ReaL & ObjNet & T$\rightarrow$I & I$\rightarrow$T & T$\rightarrow$I & I$\rightarrow$T & T$\rightarrow$I & I$\rightarrow$T & T$\rightarrow$I & I$\rightarrow$T & T$\rightarrow$I & I$\rightarrow$T \\
\midrule
\multirow[c]{12}{*}{B/16} & 64 & SigLIP 2 (NaF.) & 71.2 & 63.2 & 78.3 & 62.1 & 43.6 & 60.4 & 30.4 & 57.5 & 3.4 & 6.4 & 5.2 & 4.0 & 6.4 & 11.0 \\
\arrayrulecolor{lightgray}\cline{2-17}
 & 144 & SigLIP 2 (NaF.) & 76.2 & 69.4 & 82.9 & 70.2 & 49.0 & 65.7 & 36.5 & 65.8 & 5.7 & 10.3 & 13.5 & 11.8 & 13.9 & 25.4 \\
\arrayrulecolor{lightgray}\cline{2-17}
 & 196 & SigLIP 2 & 78.2 & 71.4 & 84.8 & 73.6 & 52.1 & 68.9 & 38.9 & 68.0 & 5.5 & 9.5 & 13.3 & 10.9 & 10.8 & 18.7 \\
\arrayrulecolor{lightgray}\cline{2-17}
 & \multirow[c]{2}{*}{256} & SigLIP 2 & 79.1 & 72.5 & 85.4 & 74.5 & 53.2 & 69.7 & 40.5 & 69.4 & 6.1 & 9.8 & 17.1 & 14.2 & 12.9 & 22.9 \\
 &  & SigLIP 2 (NaF.) & 78.5 & 71.9 & 84.6 & 74.6 & 51.1 & 67.3 & 39.5 & 69.0 & 7.4 & 12.9 & 19.7 & 17.1 & 14.8 & 26.6 \\
\arrayrulecolor{lightgray}\cline{2-17}
 & \multirow[c]{2}{*}{576} & SigLIP 2 & 80.6 & 73.8 & 86.2 & 77.1 & 54.6 & 71.4 & 43.6 & 73.0 & 7.5 & 12.0 & 23.3 & 19.4 & 14.1 & 24.8 \\
 &  & SigLIP 2 (NaF.) & 80.0 & 73.1 & 85.6 & 76.4 & 52.5 & 69.1 & 41.6 & 71.8 & 8.7 & 14.1 & 24.3 & 21.0 & 15.3 & 26.7 \\
\arrayrulecolor{lightgray}\cline{2-17}
 & 676 & SigLIP 2 (NaF.) & 80.1 & 73.5 & 85.7 & 76.5 & 52.9 & 68.6 & 41.8 & 73.0 & 8.8 & 13.9 & 24.3 & 21.4 & 15.2 & 26.2 \\
\arrayrulecolor{lightgray}\cline{2-17}
 & 784 & SigLIP 2 (NaF.) & 80.2 & 73.5 & 85.9 & 76.9 & 53.1 & 68.8 & 42.5 & 72.9 & 8.7 & 14.0 & 24.8 & 21.5 & 15.2 & 26.4 \\
\arrayrulecolor{lightgray}\cline{2-17}
 & 900 & SigLIP 2 (NaF.) & 80.3 & 73.6 & 85.9 & 76.6 & 52.9 & 69.2 & 42.3 & 72.6 & 8.6 & 15.0 & 24.8 & 21.6 & 15.0 & 25.8 \\
\arrayrulecolor{lightgray}\cline{2-17}
 & \multirow[c]{2}{*}{1024} & SigLIP 2 & 81.2 & 74.5 & 86.7 & 77.8 & 55.2 & 71.2 & 44.7 & 74.7 & 8.1 & 14.6 & 25.2 & 20.7 & 14.5 & 25.3 \\
 &  & SigLIP 2 (NaF.) & 80.4 & 73.5 & 85.9 & 76.6 & 52.9 & 68.9 & 42.5 & 73.2 & 9.1 & 14.4 & 25.1 & 21.5 & 14.9 & 26.4 \\
\arrayrulecolor{black}\cline{1-17} 
\multirow[c]{11}{*}{So/16} & 64 & SigLIP 2 (NaF.) & 78.5 & 71.0 & 84.2 & 73.8 & 49.6 & 67.4 & 37.0 & 65.5 & 5.6 & 10.3 & 11.8 & 10.9 & 12.1 & 21.4 \\
\arrayrulecolor{lightgray}\cline{2-17}
 & 144 & SigLIP 2 (NaF.) & 81.8 & 75.2 & 86.7 & 79.8 & 53.4 & 70.4 & 42.8 & 71.0 & 8.0 & 14.6 & 22.2 & 23.1 & 17.1 & 29.0 \\
\arrayrulecolor{lightgray}\cline{2-17}
 & \multirow[c]{2}{*}{256} & SigLIP 2 & 83.4 & 77.8 & 87.7 & 84.8 & 55.4 & 71.5 & 44.8 & 72.9 & 7.9 & 13.9 & 29.7 & 28.8 & 17.4 & 28.7 \\
 &  & SigLIP 2 (NaF.) & 83.5 & 77.5 & 87.7 & 83.8 & 55.1 & 71.2 & 44.9 & 73.6 & 9.2 & 15.7 & 29.8 & 29.2 & 17.5 & 29.2 \\
\arrayrulecolor{lightgray}\cline{2-17}
 & \multirow[c]{2}{*}{576} & SigLIP 2 & 84.1 & 78.4 & 88.1 & 85.8 & 56.0 & 71.2 & 47.0 & 74.9 & 9.7 & 16.3 & 34.5 & 32.4 & 17.8 & 28.0 \\
 &  & SigLIP 2 (NaF.) & 84.1 & 78.6 & 88.0 & 85.7 & 55.9 & 71.4 & 46.5 & 75.1 & 11.3 & 18.4 & 32.9 & 32.0 & 17.7 & 28.8 \\
\arrayrulecolor{lightgray}\cline{2-17}
 & 676 & SigLIP 2 (NaF.) & 84.2 & 78.5 & 88.0 & 85.7 & 55.8 & 71.7 & 46.9 & 74.9 & 11.3 & 18.5 & 33.3 & 32.2 & 17.7 & 29.8 \\
\arrayrulecolor{lightgray}\cline{2-17}
 & 784 & SigLIP 2 (NaF.) & 84.3 & 78.6 & 88.0 & 85.9 & 55.9 & 71.3 & 46.7 & 74.9 & 11.5 & 18.5 & 33.0 & 32.3 & 17.6 & 29.5 \\
\arrayrulecolor{lightgray}\cline{2-17}
 & 900 & SigLIP 2 (NaF.) & 84.3 & 78.6 & 88.1 & 85.8 & 55.8 & 71.2 & 46.8 & 75.4 & 11.7 & 18.5 & 32.9 & 32.5 & 17.7 & 29.4 \\
\arrayrulecolor{lightgray}\cline{2-17}
 & \multirow[c]{2}{*}{1024} & SigLIP 2 & 84.3 & 79.1 & 88.1 & 86.2 & 56.0 & 71.3 & 47.3 & 76.0 & 10.3 & 18.3 & 35.9 & 33.5 & 17.9 & 28.1 \\
 &  & SigLIP 2 (NaF.) & 84.4 & 78.8 & 88.1 & 85.8 & 55.8 & 71.0 & 46.9 & 74.9 & 11.7 & 18.4 & 32.6 & 32.4 & 17.8 & 29.4 \\
\arrayrulecolor{black}
\bottomrule
\end{tabular}
  \caption{Comparing the NaFlex (supporting native aspect ratio and variable sequence length (Seq.)) and the standard square-input SigLIP variants which use a separate checkpoint per sequence length. Numerical data corresponding to the plots in Fig.~\ref{fig:naflex}. TC: TextCaps, HT: HierText, SC: SciCap, S2W: Screen2Words.}
  \label{tbl:app_naflex}
\end{table*}
\vspace*{\fill}

\clearpage

\section{Full cultural diversity and fairness results}\label{app:cultural_diversity}
\vspace*{\fill}
\begin{table*}[h]
  \centering
  \footnotesize
  \setlength{\tabcolsep}{0.6em}
\begin{tabular}{lclcccccc}
\toprule
 &  &  & \multicolumn{3}{c}{10-shot} & \multicolumn{3}{c}{0-shot} \\
 \cmidrule(lr){4-6} \cmidrule(lr){7-9}
ViT & Res. & Model & Dollar Street & GeoDE (country) & GeoDE (region) & Dollar Street & GLDv2 & GeoDE \\
\midrule
B/32 & 256 & SigLIP 2 & 13.1 & 13.9 & 29.3 & 50.5 & 44.7 & 90.6 \\
\arrayrulecolor{black}\cline{1-9}
\multirow[c]{8}{*}{B/16} & \multirow[c]{2}{*}{224} & SigLIP & 13.8 & 12.7 & 27.3 & 50.1 & 48.5 & 92.4 \\
 &  & SigLIP 2 & 16.2 & 20.0 & 34.9 & 53.4 & 50.8 & 92.9 \\
\arrayrulecolor{lightgray}\cline{2-9}
 & \multirow[c]{2}{*}{256} & SigLIP & 15.0 & 13.3 & 29.3 & 50.3 & 47.7 & 92.8 \\
 &  & SigLIP 2 & 17.7 & 22.7 & 36.3 & 54.2 & 52.5 & 93.3 \\
\arrayrulecolor{lightgray}\cline{2-9}
 & \multirow[c]{2}{*}{384} & SigLIP & 16.1 & 16.4 & 31.5 & 51.5 & 51.9 & 93.6 \\
 &  & SigLIP 2 & 19.8 & 25.6 & 41.4 & 54.8 & 55.2 & 93.9 \\
\arrayrulecolor{lightgray}\cline{2-9}
 & \multirow[c]{2}{*}{512} & SigLIP & 16.6 & 17.7 & 32.3 & 51.3 & 53.1 & 94.1 \\
 &  & SigLIP 2 & 21.7 & 28.2 & 43.1 & 54.9 & 57.6 & 94.2 \\
\arrayrulecolor{black}\cline{1-9} 
\multirow[c]{5}{*}{L/16} & \multirow[c]{2}{*}{256} & SigLIP & 18.8 & 22.1 & 36.2 & 52.1 & 56.7 & 93.6 \\
 &  & SigLIP 2 & 26.8 & 34.5 & 44.4 & 55.2 & 64.5 & 94.9 \\
\arrayrulecolor{lightgray}\cline{2-9}
 & \multirow[c]{2}{*}{384} & SigLIP & 22.8 & 26.0 & 41.7 & 52.9 & 60.5 & 94.3 \\
 &  & SigLIP 2 & 30.4 & 39.3 & 48.0 & 55.4 & 66.1 & 95.1 \\
\arrayrulecolor{lightgray}\cline{2-9}
 & 512 & SigLIP 2 & 32.5 & 42.5 & 50.6 & 55.2 & 67.6 & 95.3 \\
\arrayrulecolor{black}\cline{1-9} 
\multirow[c]{4}{*}{So400m/14} & \multirow[c]{2}{*}{224} & SigLIP & 26.6 & 31.9 & 45.8 & 55.1 & 74.1 & 94.7 \\
 &  & SigLIP 2 & 31.9 & 38.1 & 49.1 & 55.4 & 65.6 & 94.8 \\
\arrayrulecolor{lightgray}\cline{2-9}
 & \multirow[c]{2}{*}{384} & SigLIP & 32.1 & 36.5 & 51.6 & 56.3 & 71.7 & 94.9 \\
 &  & SigLIP 2 & 38.3 & 45.2 & 56.1 & 56.6 & 68.6 & 95.2 \\
\arrayrulecolor{black}\cline{1-9} 
\multirow[c]{4}{*}{So400m/16} & \multirow[c]{2}{*}{256} & SigLIP 2 & 33.2 & 39.8 & 50.9 & 55.8 & 66.7 & 95.0 \\
 &  & mSigLIP & 27.1 & 33.3 & 48.5 & 54.2 & 57.5 & 94.3 \\
\arrayrulecolor{lightgray}\cline{2-9}
 & 384 & SigLIP 2 & 38.2 & 44.1 & 54.4 & 56.5 & 67.8 & 95.3 \\
\arrayrulecolor{lightgray}\cline{2-9}
 & 512 & SigLIP 2 & 40.8 & 47.6 & 58.6 & 56.6 & 69.2 & 95.3 \\
\arrayrulecolor{black}\cline{1-9} 
\multirow[c]{2}{*}{g-opt/16} & 256 & SigLIP 2 & 37.6 & 46.6 & 54.0 & 56.9 & 71.2 & 95.4 \\
\arrayrulecolor{lightgray}\cline{2-9}
 & 384 & SigLIP 2 & 44.5 & 52.0 & 58.7 & 57.2 & 72.2 & 95.7 \\
\arrayrulecolor{black}
\bottomrule
\end{tabular}
  \caption{10-shot and 0-shot accuracy for geographically diverse object classification tasks (Dollar Street, GeoDE), as well as geolocalization (GeoDE country/region) and landmark localization (GLDv2) tasks. SigLIP 2 consistently outperforms SigLIP on most benchmarks.}
  \label{tbl:app_cultural_diversity}
\end{table*}
\vspace*{\fill}

\begin{table*}
  \centering
  \footnotesize
  \begin{tabular}{lclcc}
\toprule
ViT & Res. & Model & Disparity & Rep. bias \\
\midrule
B/32 & 256 & SigLIP 2 & 33.3 & 16.6 \\
\arrayrulecolor{black}\cline{1-5} 
\multirow[c]{8}{*}{B/16} & \multirow[c]{2}{*}{224} & SigLIP & 31.2 & 36.6 \\
 &  & SigLIP 2 & 31.0 & 17.2 \\
\arrayrulecolor{lightgray}\cline{2-5}
 & \multirow[c]{2}{*}{256} & SigLIP & 30.2 & 35.6 \\
 &  & SigLIP 2 & 29.7 & 19.4 \\
\arrayrulecolor{lightgray}\cline{2-5}
 & \multirow[c]{2}{*}{384} & SigLIP & 30.9 & 35.8 \\
 &  & SigLIP 2 & 30.6 & 18.0 \\
\arrayrulecolor{lightgray}\cline{2-5}
 & \multirow[c]{2}{*}{512} & SigLIP & 31.5 & 35.4 \\
 &  & SigLIP 2 & 30.8 & 20.0 \\
\arrayrulecolor{black}\cline{1-5} 
\multirow[c]{5}{*}{L/16} & \multirow[c]{2}{*}{256} & SigLIP & 32.0 & 35.5 \\
 &  & SigLIP 2 & 31.1 & 7.3 \\
\arrayrulecolor{lightgray}\cline{2-5}
 & \multirow[c]{2}{*}{384} & SigLIP & 32.0 & 34.8 \\
 &  & SigLIP 2 & 30.4 & 6.6 \\
\arrayrulecolor{lightgray}\cline{2-5}
 & 512 & SigLIP 2 & 29.2 & 6.8 \\
\arrayrulecolor{black}\cline{1-5} 
\multirow[c]{4}{*}{So400m/14} & \multirow[c]{2}{*}{224} & SigLIP & 30.5 & 33.3 \\
 &  & SigLIP 2 & 29.7 & 7.4 \\
\arrayrulecolor{lightgray}\cline{2-5}
 & \multirow[c]{2}{*}{384} & SigLIP & 29.2 & 33.9 \\
 &  & SigLIP 2 & 28.1 & 7.5 \\
\arrayrulecolor{black}\cline{1-5} 
\multirow[c]{4}{*}{So400m/16} & \multirow[c]{2}{*}{256} & SigLIP 2 & 28.4 & 7.2 \\
 &  & mSigLIP & 31.6 & 37.3 \\
\arrayrulecolor{lightgray}\cline{2-5}
 & 384 & SigLIP 2 & 29.0 & 11.0 \\
\arrayrulecolor{lightgray}\cline{2-5}
 & 512 & SigLIP 2 & 28.2 & 10.8 \\
\arrayrulecolor{black}\cline{1-5} 
\multirow[c]{2}{*}{g-opt/16} & 256 & SigLIP 2 & 28.1 & 7.9 \\
\arrayrulecolor{lightgray}\cline{2-5}
 & 384 & SigLIP 2 & 28.3 & 4.9 \\
\arrayrulecolor{black}
\bottomrule
\end{tabular}
  \caption{Disparity: Corresponds to the maximum difference in 0-shot accuracy on Dollar Street when disaggregating the accuracy by income level: We observe that SigLIP 2 slightly reduces the performance disparity. Rep. bias: Representation bias; lower values are better. SigLIP2, which is trained on de-biased data, exhibits significantly reduced representation bias than its predecessor. In addition, larger models are better than smaller models, in agreement with the earlier findings in~\cite{alabdulmohsin2024clip}.}
  \label{tbl:app_rb}
\end{table*}

\end{document}